\documentclass[10pt,twocolumn,letterpaper]{article}

\usepackage{cite}

\usepackage{cvpr}
\usepackage{times}
\usepackage{epsfig}
\usepackage{graphicx}
\usepackage{amsmath}
\usepackage{amssymb}

\newcommand{\figdir}{figures}
\def\swone{0.98\linewidth}

\def\swtwo{0.49\linewidth}
\def\swthree{0.33\linewidth}
\def\swfour{0.24\linewidth}
\def\swfive{0.19\linewidth}
\def\swsix{0.164\linewidth}

\newcommand{\reffig}[1]{Figure. \ref{#1}}

\newcommand{\refsec}[1]{Section. \ref{#1}}

\long\def\ignorethis#1{}

\usepackage[pagebackref=true,breaklinks=true,letterpaper=true,colorlinks,citecolor=blue,linkcolor=blue,bookmarks=false]{hyperref}

 \cvprfinalcopy 

\def\cvprPaperID{1464} 

\ifcvprfinal\fi
\begin{document}

%
\title{Learning Fully Convolutional Networks for Iterative Non-blind Deconvolution}

\author{Jiawei Zhang$^{1,3}$\\
\and
Jinshan Pan$^{2,3}$\\
\and
Wei-Sheng Lai$^3$\\
\and
Rynson Lau$^1$\\
\and
Ming-Hsuan Yang$^3$\\
\and
Department of Computer Science, City University of Hong Kong$^1$\\
School of Mathematical Sciences, Dalian University of Technology$^2$\\
Electrical Engineering and Computer Science, University of California, Merced$^3$\\
}

\maketitle

\begin{abstract}
In this paper, we propose a fully convolutional networks for iterative non-blind deconvolution
%
We decompose the non-blind deconvolution problem into image denoising and image deconvolution.
We train a FCNN to remove noises in the gradient domain and use the learned gradients to guide the image deconvolution step.
In contrast to the existing deep neural network based methods,
we iteratively deconvolve the blurred images in a multi-stage framework.
The proposed method is able to learn an adaptive image prior, which keeps both local (details) and global (structures) information.
%
%
Both quantitative and qualitative evaluations on benchmark datasets demonstrate that
the proposed method performs favorably against state-of-the-art algorithms in terms of quality and speed.
\end{abstract}

\section{Introduction}
Single image non-blind deconvolution aims to recover a sharp latent image given a blurred image and the blur kernel.
%
The community has made active research effort on this classical problem with the last decade.
%
Assuming the camera motion is spatially invariant, a blurred image $y$ can be modeled as a convolution between a blur kernel $k$ and a latent image $x$:
\begin{equation} \label{eq:blur}
	y=k*x+n,
\end{equation}
where $n$ is additive noise and $*$ is the convolution operator.
In non-blind deconvolution, we solve $x$ from $y$ and $k$, which is an ill-posed problem since the noise is unknown.

%
%
%
%
Conventional approaches, such as the Richardson-Lucy deconvolution~\cite{RL} and the Wiener filter~\cite{wiener1949extrapolation}, suffer from serious ringing artifacts and are less effective to deal with large motion and outliers.
Several methods focus on developing effective image priors for image restoration, including Hyper-Laplacian priors~\cite{krishnan2009fast,levin2009understanding},
non-local means~\cite{nonlocal/means/cvpr06}, field of experts~\cite{DBLP:journals/pami/SchmidtJNRR16,DBLP:conf/cvpr/SchmidtSR11,schmidt2013discriminative,DBLP:conf/cvpr/RothB05},
patch-based priors~\cite{zoran2011learning,sun/goodimageprior/eccv14} and shrinkage fields~\cite{schmidt2014shrinkage}.
However, these image priors are heavily based on the empirical statistics of natural images.   
%
%
In addition, these image priors typically lead to highly non-convex optimization problems.
Most of the aforementioned methods need expensive computational costs to obtain state-of-the-art deblurred results.
%

%
%
%
Recently, the deep neural network has been applied to image restoration~\cite{schuler2013machine,xu2014deep}.
%
%
However, these methods need to re-train the network for different blur kernels, which is not practical in real-world scenarios.

Different from existing methods, we propose a FCNN for iterative non-blind deconvolution, 
which is able to automatically learn effective image prior and does not need to re-train the network for different blur kernels.
The proposed method decomposes the non-blind deconvolution into two steps: image denoising and image deconvolution.
In the image denoising step, we train a FCNN to remove noises and outliers in the \emph{gradient} domain.
The learned image gradients are treated as image priors to guide the image deconvolution.
In the image deconvolution step, we concatenate a deconvolution module at the end of the FCNN to remove the blur from input images.
%
We cascade the FCNN into a multi-stage architecture to deconvolve blurred images iteratively.
%
The proposed FCNN adaptively learns effective image priors to preserve image details and structures.
%
%
%
In order to effectively suppress ringing artifacts and noises in the smooth regions, we propose to optimize the FCNN with a robust $L_1$ loss function instead of a commonly used $L_2$ loss function.
%
In addition, we optimize the hyper-parameters in the deconvolution modules.
Extensive evaluation on benchmark datasets demonstrates that the proposed method performs favorably against state-of-the-art algorithms in terms of quality and speed.

\begin{figure*}[!htb]
	\begin{center}
		\includegraphics[width=\swone]{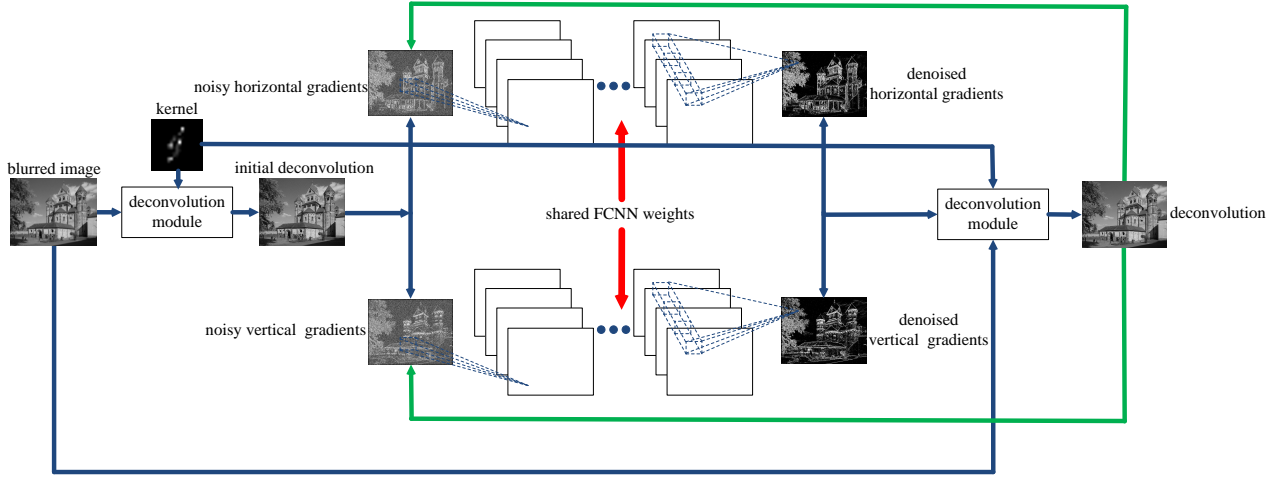}
	\end{center}
	\caption{Network structure.
Our network first deconvolves blurry input images by the deconvolution module and then performs convolutions to the vertical and horizontal gradients to generate the results with fewer noises. Finally, the deconvolution module is applied to the denoised gradients to generate the clear images. See text for more details.
}
	\label{fig:network}
\end{figure*}

\section{Related Work}
Non-blind deconvolution has been studied extensively and numerous algorithms have been proposed.
In this section we, discuss the most relevant algorithms and put
this work in proper context.

Since non-blind deblurring is an ill-posed problem, it requires some assumptions or prior knowledge to constrain the solution space.
%
%
The early approaches, \eg, Wiener deconvolution \cite{wiener1949extrapolation}, assume that the value of every pixel should follow Gaussian distribution.
%
However, this assumption does not hold for natural images as the distribution of real-world image gradient is heavy-tailed.
%
%
To develop an image prior that fits the heavy-tailed distribution of natural images, the Hyper-Laplacian prior is proposed~\cite{levin2009understanding}.
As solving the image restoration with the Hyper-Laplacian prior is time-consuming, Krishnan and Fergus~\cite{krishnan2009fast} propose
an efficient algorithm based on a half-quadratic splitting method.

To learn good priors for image restoration, Roth and Black~\cite{DBLP:conf/cvpr/RothB05} learns a group of field of experts (FOEs) to fit the heavy-tailed distribution of natural images.
The FOE framework is further extended by~\cite{schmidt2013discriminative, schmidt2014shrinkage}. However, methods with field of expert usually lead to complex optimization problems.
Solving these problems are usually time-consuming.


The Gaussian Mixture Model (GMM) has also been developed to fit the heavy-tailed distribution of natural image gradient. Fergus~\etal~\cite{FergusACM2006} use a mixture of Gaussians to learn
an image gradient prior via variational Bayesian inference.
Zoran and Weiss \cite{zoran2011learning} analyze the image priors in image restoration and propose a patch based prior based on GMM.
This work is further extended by Sun~\etal~\cite{sun/goodimageprior/eccv14}.
Although good results have been achieved, solving these methods needs heavy computation loads.
%



%
%

Recently deep learning has been used for low-level image processing such as denoising \cite{eigen2013restoring, burger2012image, xie2012image, jain2009natural},
super-resolution \cite{dong2014learning, wang2015deep, riegler2016deep, kim2015accurate, riegler2016atgv, kim2015deeply, yu2016ultra}
and edge-preserving filtering \cite{xu2015deep, li2016deep, liu2016learning}.
%

%
For non-blind deblurring, Schuler~\etal~\cite{schuler2013machine} develop a multi-layer perceptron (MLP) approach to remove noise and artifacts which are produced by the deconvolution process.
Xu~\etal~\cite{xu2014deep} use a deep CNN to restore images with outliers. This method uses singular value decomposition (SVD) to reduce the number of parameters in the network.
However, it needs to fine-tune the network for every kernel as it uses SVD of the pseudo inverse kernel as the network initialization.
Different from existing CNN-based method, we develop an effective FCNN for iterative non-blind deconvolution.
We cascade the FCNN into a multi-stage architecture to deconvolve blurred images to iteratively preserve
the details of the restored images.
Moreover, our method does not retain model for each blur kernel.
It is much faster than the existing image restoration methods.


\section{Proposed Algorithm}
In this section, we present our algorithm to learn effective image prior for non-blind image deconvolution.
We first review half-quadratic optimization in image restoration and then introduce our method.
\subsection{Motivation}
Half-quadratic splitting framework has been widely used in non-blind deblurring methods \cite{krishnan2009fast, zoran2011learning, schmidt2013discriminative, schmidt2014shrinkage}.
In this section, we first review this method in image restoration and then motivate our method.
The conventional model of image restoration is
\begin{equation}
\label{eq: deblur-model}
\min_{x}\frac{\lambda}{2}\|y - x*k \|_2^2 + \sum_{l=h,w}\rho(p_l*x),
\end{equation}
where $p_h,p_w$ are horizontal and vertical gradient operators, $\rho(\cdot)$ is the regularization of image gradient of $x$.
With the half-quadratic splitting method, model~\eqref{eq: deblur-model} can be reformulated as
\begin{equation}
\label{eq: deblur-model-hq}
\min_{x, z}\frac{\lambda}{2}\|y - x*k \|_2^2 + \beta \sum_{l=h,w} \|z_l- p_l*x\|_2^2 + \rho(z_l),
\end{equation}
where $z_l$ is an auxiliary variable and $\beta$ is a weight.
The half-quadratic optimization with respect to~\eqref{eq: deblur-model-hq} is to alternatively solve
\begin{equation}
\label{eq: deblur-model-hq-aux}
\min_{z} \beta \sum_{l=h,w} \|z_l- p_l*x\|_2^2 + \rho(z_l),
\end{equation}
and
\begin{equation}
\label{eq: deblur-model-hq-x}
\min_{x} \frac{\lambda}{2}\|y - x*k \|_2^2 + \beta \sum_{l=h,w} \|z_l- p_l*x\|_2^2,
\end{equation}

We note that~\eqref{eq: deblur-model-hq-aux} is actually a denoising problem while~\eqref{eq: deblur-model-hq-x} is a deconvolution with respect to $x$.
If the solution of $z_l$ is obtain, the clear image can be efficiently solved by fast Fourier transform (FFT),
\begin{equation}
\label{eq: solution-of-x}
x=\mathcal{F}^{-1}\left(\frac{\gamma\overline{\mathcal{F}(k)} \mathcal{F}(y) +
	\sum_{l=h,w}\overline{\mathcal{F}(p_l)}\mathcal{F}(z_l)} {\gamma\overline{\mathcal{F}(k)} \mathcal{F}(k) + \sum_{l=h,w}\overline{\mathcal{F}(p_l)} \mathcal{F}(p_l)} \right)
\end{equation}
where $\mathcal{F}(\cdot)$ and $\mathcal{F}^{-1}(\cdot)$ denote the Fourier transform and its inverse transform, respectively,
$\overline{\mathcal{F}(\cdot)}$ is the complex conjugate operator
and $\gamma=\frac{\lambda}{2\beta}$ is the hyperparameter in deconvolution.

We note that the main problem is how to define a good image prior to~\eqref{eq: deblur-model-hq-aux}.
%
%
In the following, we propose an effective algorithm based on FCNN to learn an effective image prior for~\eqref{eq: deblur-model-hq-aux}.

\renewcommand{\tabcolsep}{5pt}
\begin{table}[!t]
	\centering
	\caption{The architecture of FCNN in one iteration.}
	\vspace{3mm}
	\label{table:denoiseFCNN}
\begin{tabular}{|c|c|c|c|c|}
	\hline
	name  & kernel size & stride & pad & kernel number \\ \hline
	conv1 & $5\times5$         & 1      & 2   & 64            \\ \hline
	conv2 & $3\times3$        & 1      & 1   & 64            \\ \hline
	conv3 & $3\times3$       & 1      & 1   & 64            \\ \hline
	conv4 & $3\times3$         & 1      & 1   & 64            \\ \hline
	conv5 & $3\times3$         & 1      & 1   & 64            \\ \hline
	conv6 & $3\times3$         & 1      & 1   & 1             \\ \hline
\end{tabular}
\end{table}

\subsection{Network Architecture}
%

The proposed network architecture for non-blind deconvolution is shown in Figure~\ref{fig:network}.
The input of our network includes blurry image and the corresponding blur kernel.
The proposed network first applies the deconvolution operation on the blurry images via a deconvolution module and
then performs convolutions to the vertical and horizontal gradients to generate the results with fewer noises.
The denoised image gradients are treated as image priors to guide the image deconvolution in the next iteration.
%
%
%

\begin{figure*}
	\begin{center}
		\vspace{-1mm}\begin{tabular}{cccc}
			\vspace{-1mm}\includegraphics[width=\swfour]{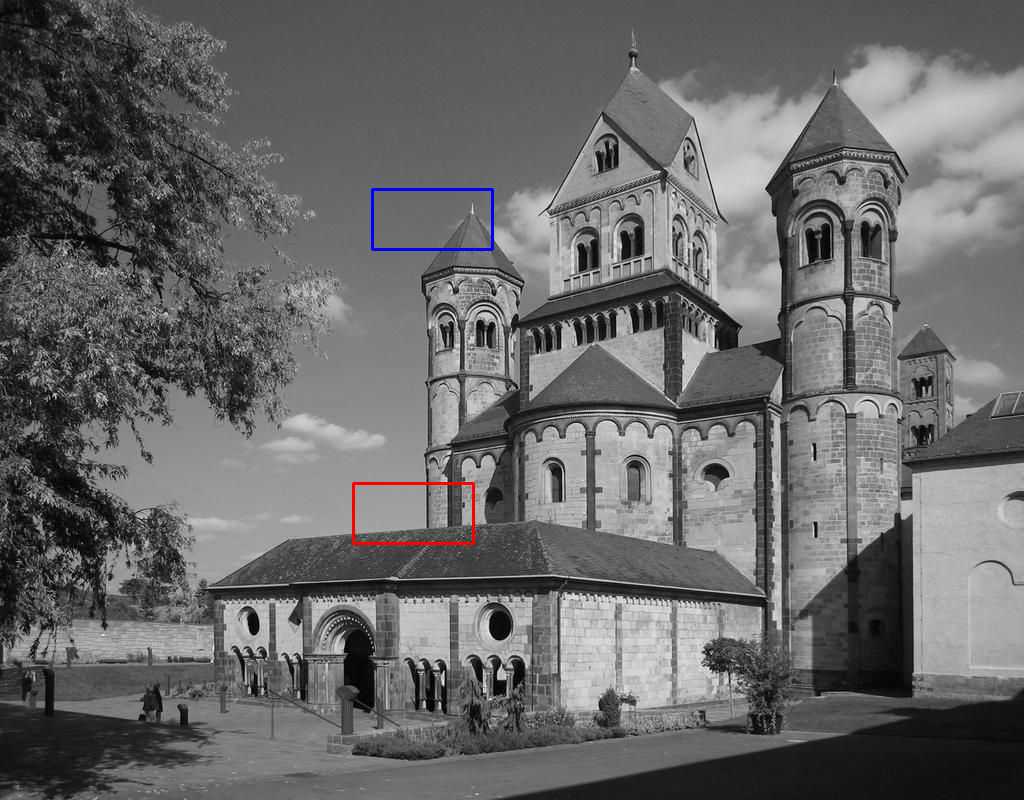}&
			\includegraphics[width=\swfour]{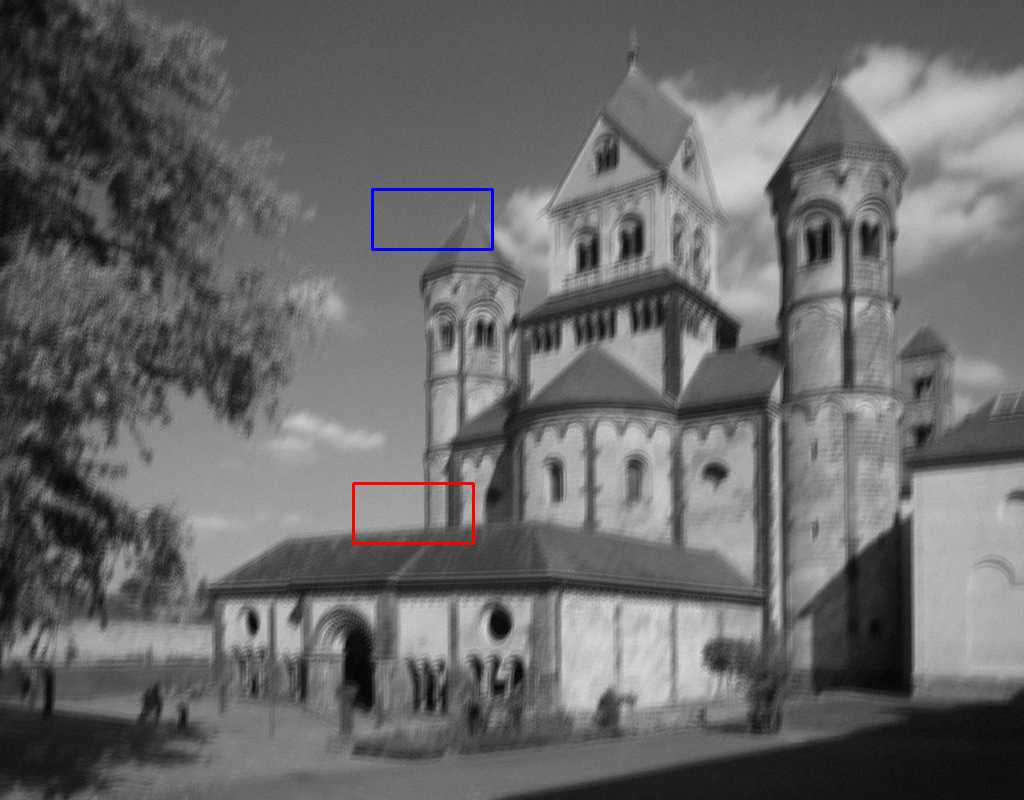}&
			\includegraphics[width=\swfour]{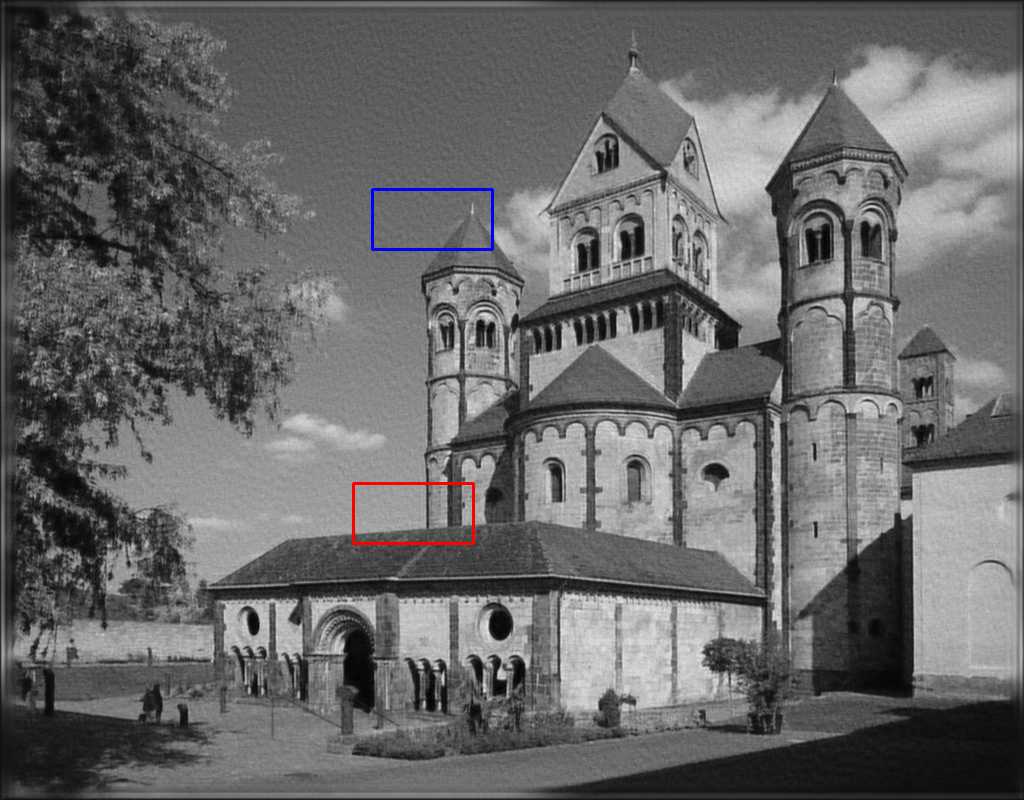}&
			\includegraphics[width=\swfour]{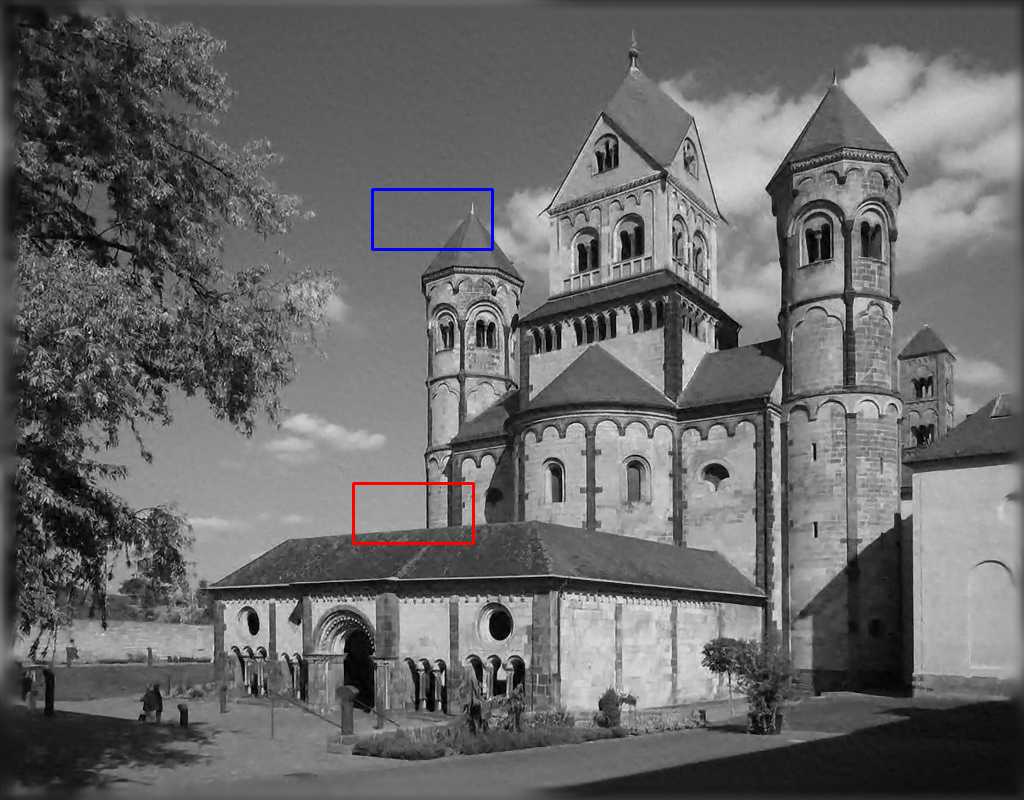}\\
			\footnotesize{(a) Clean image}&\footnotesize{(b) Blurry image}&\footnotesize{(c) Intensity domain output}&\footnotesize{(d) Gradient domain output}\\
		\end{tabular}
		\vspace{-1mm}\begin{tabular}{ccccc}
			\vspace{-1mm}\includegraphics[width=\swfive]{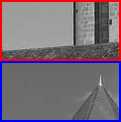}&
			\includegraphics[width=\swfive]{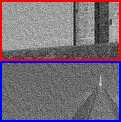}&
			\includegraphics[width=\swfive]{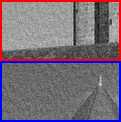}&
			\includegraphics[width=\swfive]{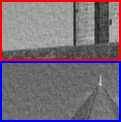}&
			\includegraphics[width=\swfive]{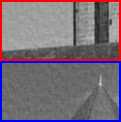}\\
			\footnotesize{(e) Local region of (a)}&\footnotesize{(f) Intensity: Initial result}&\footnotesize{(g) Intensity: 1st iteration}&\footnotesize{(h) Intensity: 2nd iteration}&\footnotesize{(i) Intensity: 3rd iteration}\\
			\vspace{-1mm}\includegraphics[width=\swfive]{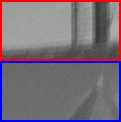}&
			\includegraphics[width=\swfive]{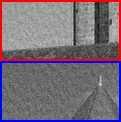}&
			\includegraphics[width=\swfive]{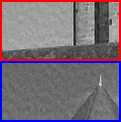}&
			\includegraphics[width=\swfive]{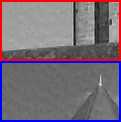}&
			\includegraphics[width=\swfive]{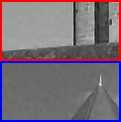}\\
			\footnotesize{(j) Local region of (b)}&\footnotesize{(k) Gradient: Initial result}&\footnotesize{(l) Gradient: 1st iteration}&\footnotesize{(m) Gradient: 2nd iteration}&\footnotesize{(n) Gradient: 3rd iteration}\\
		\end{tabular}
	\end{center}
	\caption{Visual results generation after each iteration in different domains. The clean image is shown in (a) and blurry image is shown in (b). The output results generated from the intensity and gradient domain are shown in (c) and (d), respectively. The extracted local regions of (a) and (b) are shown in (e) and (j). When the network is applied on the intensity domain we shown initial deconvolution result in (f) and refined results through different iterations from (g)-(i). When the network is applied on the gradient domain we shown initial deconvolution result in (k) and refined results through different iterations from (l)-(n). It indicates that in gradient domain our network is more effective to reduce noise through several iterations.}
	\label{fig:multiStage_deconv}
\end{figure*}

%
{\flushleft \bf{Denoising by FCNN.}}
We note that though the output of deconvolution $x$ from \eqref{eq: solution-of-x} is sharp, it usually contains noises and significant ringing artifacts (see Figure~\ref{fig:multiStage_deconv}(k)).
To solve this problem, we develop a FCNN and apply it to the vertical and horizontal gradients to remove noises and ringing artifacts.
Applying FCNN to the vertical and gradient horizontal gradients usually leads to different network weight parameters.
Similar to \cite{xu2015deep}, we transpose the vertical gradient so vertical gradient and horizontal gradient can share the weights in the training process.
Table~\ref{table:denoiseFCNN} shows the details of the proposed network.
We add a rectified linear unit (ReLU) after every convolution layer as activation function except the last one.
%
%

%
{\flushleft \bf{Deconvolution module.}}
%
%
The deconvolution module is used to restore sharp images. It is defined as~\eqref{eq: deblur-model-hq-x}.
In the proposed network, it is applied to the gradient denoising outputs from FCNN to guide the image restoration.

%

\renewcommand{\tabcolsep}{.1pt}

\subsection{Loss Function for FCNN Training}
Since it is very difficult to train the network in an end-to-end manner, we train the weights of FCNN.
That is, we first train the network weights and then fix these weights when performing the deconvolution.
After the deconvolution module, the trained network weights is then updated.
%
This training procedure is achieved by minimizing the loss function $L$:
\begin{equation}
\begin{split}
\label{eq: L1Norm}
L(\bigtriangledown_h x,\bigtriangledown_w x,x_0;\theta)=&\frac{1}{N}\sum_{i=1}^N (\|f(\bigtriangledown_h x^{(i)};\theta)-\bigtriangledown_h x_0^{(i)}\|_1\\
&+ \|f(\bigtriangledown_w x^{(i)};\theta)-\bigtriangledown_w x_0^{(i)}\|_1)
\end{split}
\end{equation}
where $f(\cdot)$ is the denoising mapping learned by FCNN, $\theta$ is the FCNN weights, $\bigtriangledown_h x= p_h*x$,  $\bigtriangledown_w x=p_w*x$, $N$ is the number of training samples in every batch, $\|\cdot\|_1$ is $L_1$ norm and $x_0$ is the ground truth image.

\subsection{Hyper-parameters Training}
In order to get optimal hyper-parameters $\gamma$ for the deconvolution module~\eqref{eq: deblur-model-hq-x},
we train them in an end-to-end manner with fixed FCNN weights. 
The hyper-parameters training process is achieved by minimizing the loss function
\begin{align} \label{eq:hyper-loss}
\mathcal{L}^{h} =\frac{1}{N} \sum_i^{N}\| x^{(i)} - x_0^{(i)}\|_1,
\end{align}
where $x$ is the output of the final deconvolution module.

As the forward propagation of deconvolution module is defined as \eqref{eq: solution-of-x},
we can get the gradient in backward propagation by
\begin{align} \label{eq:BIF}
\Delta_{z_l}
=&\mathcal{F}\left(\frac{\overline{\mathcal{F}(p_l)}\mathcal{F}^{-1}(\Delta_x)} {\gamma \overline{\mathcal{F}(k)}\mathcal{F}(k)+\sum_{l=h,w}\overline{\mathcal{F}(p_l)} \mathcal{F}(p_l)}\right),
\end{align}
where $\Delta_x = \frac{x^{(i)} - x_0^{(i)}}{| x^{(i)} - x_0^{(i)}|}$.


%

The gradient that is used to update of the hyper-parameters $\gamma$ can be written as:
\begin{align} \label{eq:BIFLambda}
\Delta_{\gamma} = \left(\frac{\mathbf{D H}-\mathbf{E G}}{(\gamma \mathbf{G}+\mathbf{H})^2}\right)^{\top}\mathbf{L}_x^h,
\end{align}
where $\mathbf{D}$, $\mathbf{H}$, $\mathbf{E}$, $\mathbf{G}$, and $\mathbf{L}_x$ denote the vector forms of $D$, $H$, $E$, $G$ and $\mathcal{L}^{h}_x$, respectively,
in which $D=\overline{\mathcal{F}(k)} \mathcal{F}(y)$, $E=\sum_{l=h,w}\overline{\mathcal{F}(p_l)} \mathcal{F}(z_l)$, $G=\overline{\mathcal{F}(k)} \mathcal{F}(k)$, $H=\sum_{l=h,w}\overline{\mathcal{F}(p_l)} \mathcal{F}(p_l)$,
and $L_x^{h} = \mathcal{F}^{-1}(\Delta_x)$.
The detailed derivations about~\eqref{eq:BIF} and~\eqref{eq:BIFLambda} are included in the supplemental material.

\section{Analysis and Discussion}
In this section, we analyze the effect of FCNN for iterative deconvolution, demonstrate why the gradient domain is used,
and validate the proposed loss function used in the proposed network.

\subsection{Effect of FCNN for iterative deconvolution} \label{sec:iterWiseFCNN}
In the proposed method, we iteratively solve the deconvolution and denoising part. That is, network parameters of FCNN are updated at each iteration.
With this manner, the high-quality results can be obtained.

%

\begin{figure}[!t]
	\begin{center}
		\begin{tabular}{cc}
			\vspace{-1mm}\includegraphics[width=\swtwo]{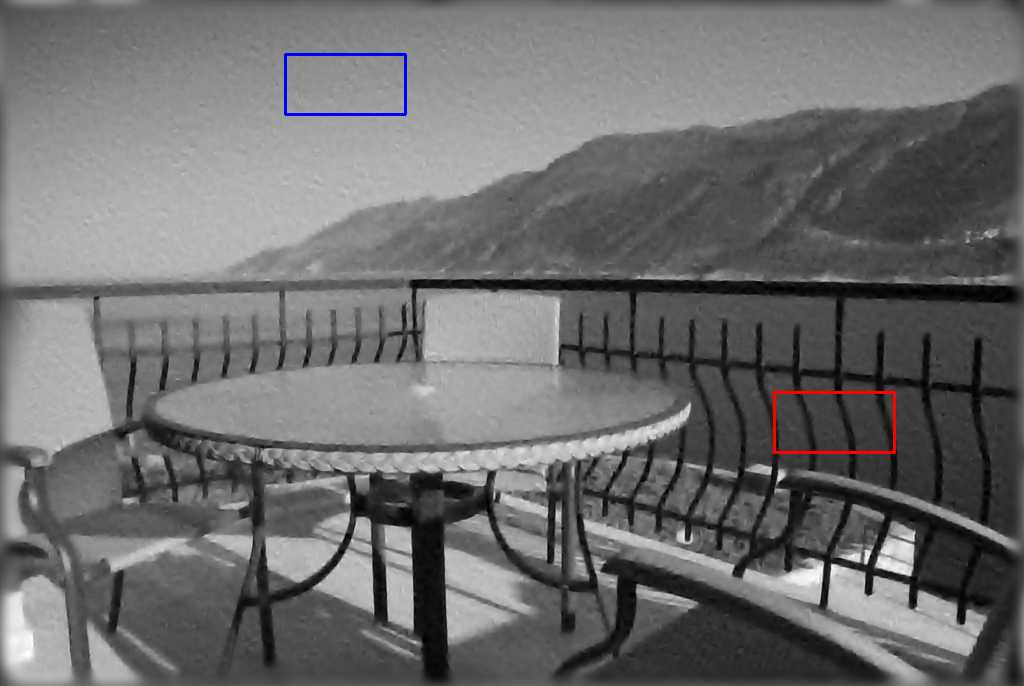} &
			\includegraphics[width=\swtwo]{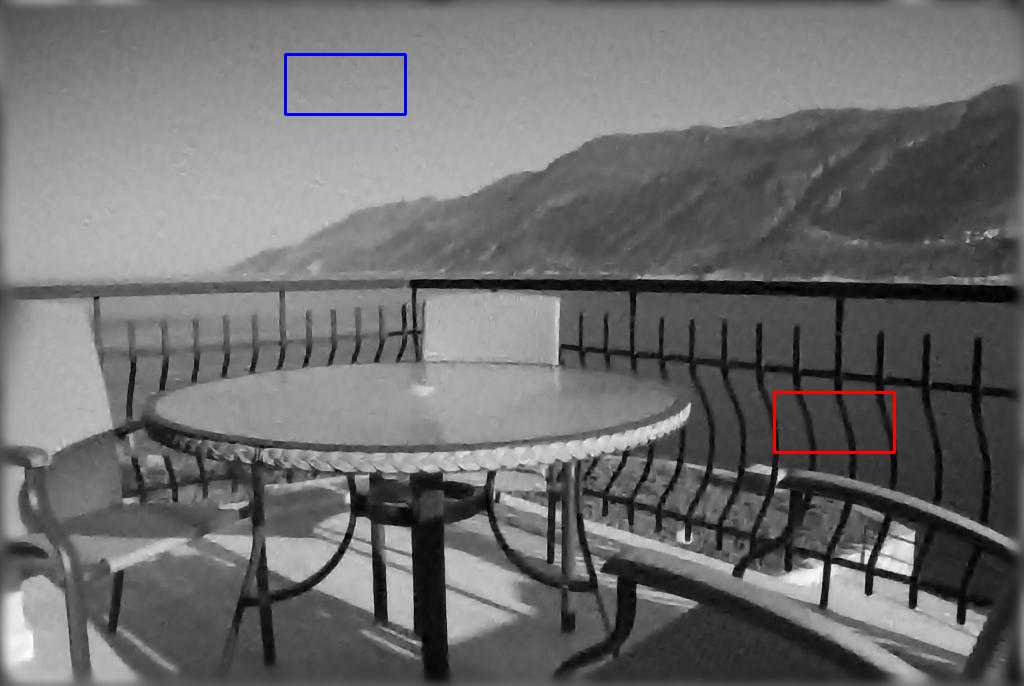} \\
			\includegraphics[width=\swtwo]{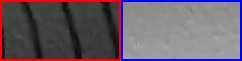} &
			\includegraphics[width=\swtwo]{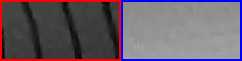} \\
			\footnotesize{(a) One iteration network} & \footnotesize{(b) Three iterations network}\\
			\footnotesize{PSNR: 29.26 dB} &\footnotesize{PSNR: 29.85 dB}\\
		\end{tabular}
	\end{center}
	\caption{The effectiveness of the FCNN for iterative deconvolution. Using only one iteration does not remove the noises. Please see \refsec{sec:iterWiseFCNN} for details.}
	\label{fig:1itervs3iter}
\end{figure}

%
Figure~\ref{fig:1itervs3iter} shows an example which demonstrates the effectiveness of the FCNN for iterative deconvolution.
As shown in Figure~\ref{fig:1itervs3iter}(a), the result generated by one iteration network contains some artifacts and has a lower PSNR value.
In contrast, these artifacts are reduced by the iterative FCNN which accordingly lead to a much clearer image with higher PSNR value (Figure~\ref{fig:1itervs3iter}(b)).
Please noted that one iteration network is different from the first iteration output of the three iterations network such as \reffig{fig:multiStage_deconv} (l).
We optimize the FCNN weights and deconvolution hyper-parameters for the network with only one iteration.
This will also apply for the one iteration network in the experiment section.

\renewcommand{\tabcolsep}{.1pt}


\begin{figure}[!t]
	\begin{center}
		\includegraphics[width=\swone]{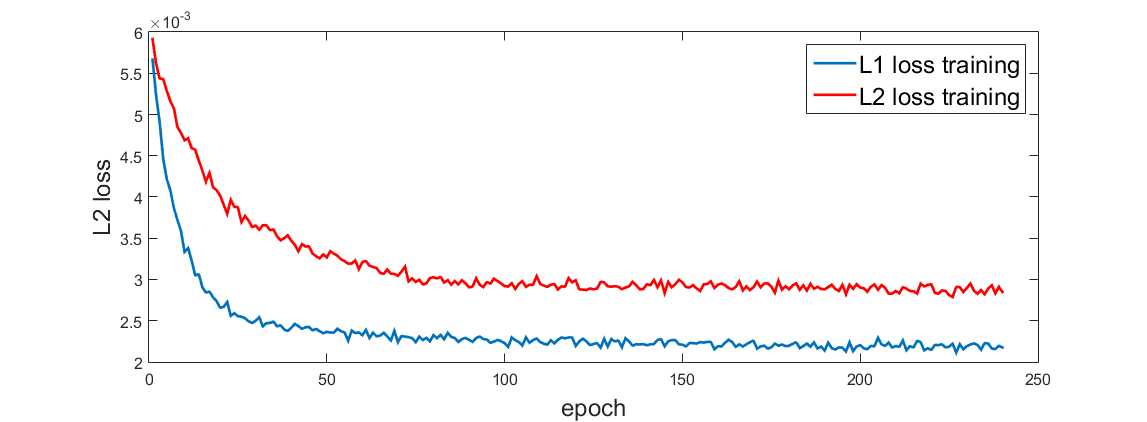}
	\end{center}
	\caption{The 1st iteration FCNN denoising gradient training loss with 1\% noises. The converged L2 norm training loss is higher than L1 norm.}
	\label{fig:L1vsL2_Loss}
\end{figure}

%


\subsection{Gradient Domain v.s. Intensity Domain}
%
The denoising part is mainly used to remove noise and ringing artifacts while keeping textures.
We note that the image gradient is able to model the details and structures of images. Thus, we train the network in image gradient domain instead of intensity domain.
We train two three iterations network where the restored results are computed based on intensity domain and gradient domain.
As shown in \reffig{fig:multiStage_deconv}, the results reconstructed from intensity domain (the second row) contains several noise and artifacts relative that from gradient domain (the third row).


\renewcommand{\tabcolsep}{.1pt}
\begin{figure}
	\begin{center}
		\begin{tabular}{cc}
			\begin{minipage}[b]{0.01\linewidth}
				\centering
				$\vcenter{\rotatebox{90}{L2 loss}}$
			\end{minipage}
			\vspace{-1mm}\begin{minipage}[t]{0.99\linewidth}
				\centering
				\begin{tabular}{cc}
					\includegraphics[width=0.45\linewidth]{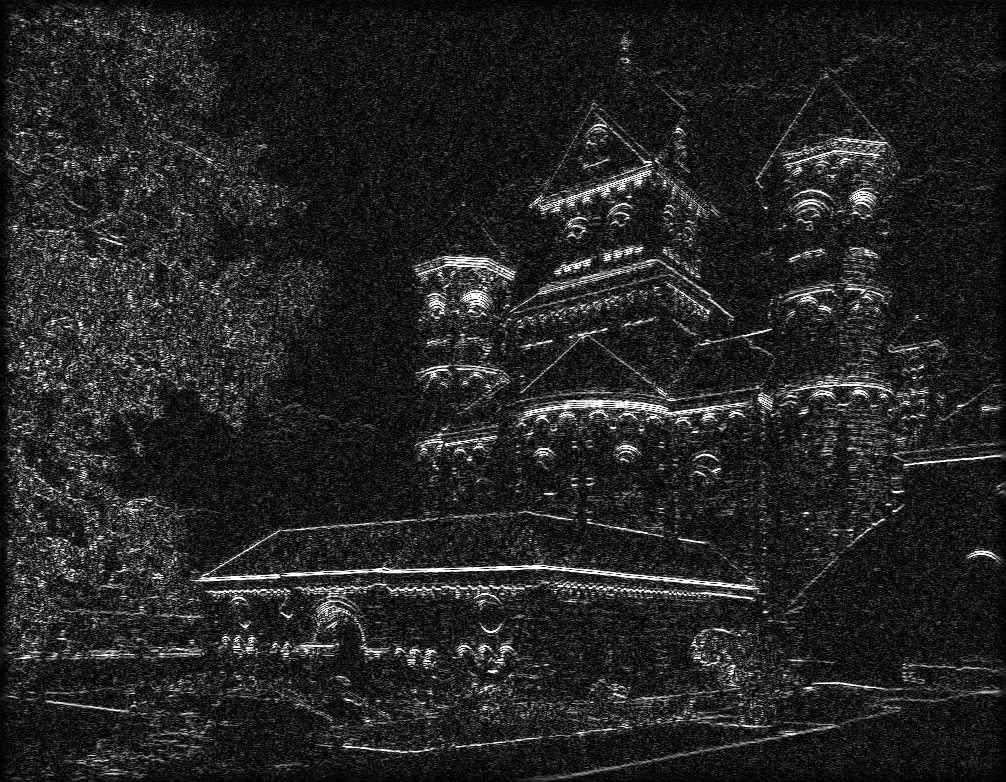}&
					\includegraphics[width=0.45\linewidth]{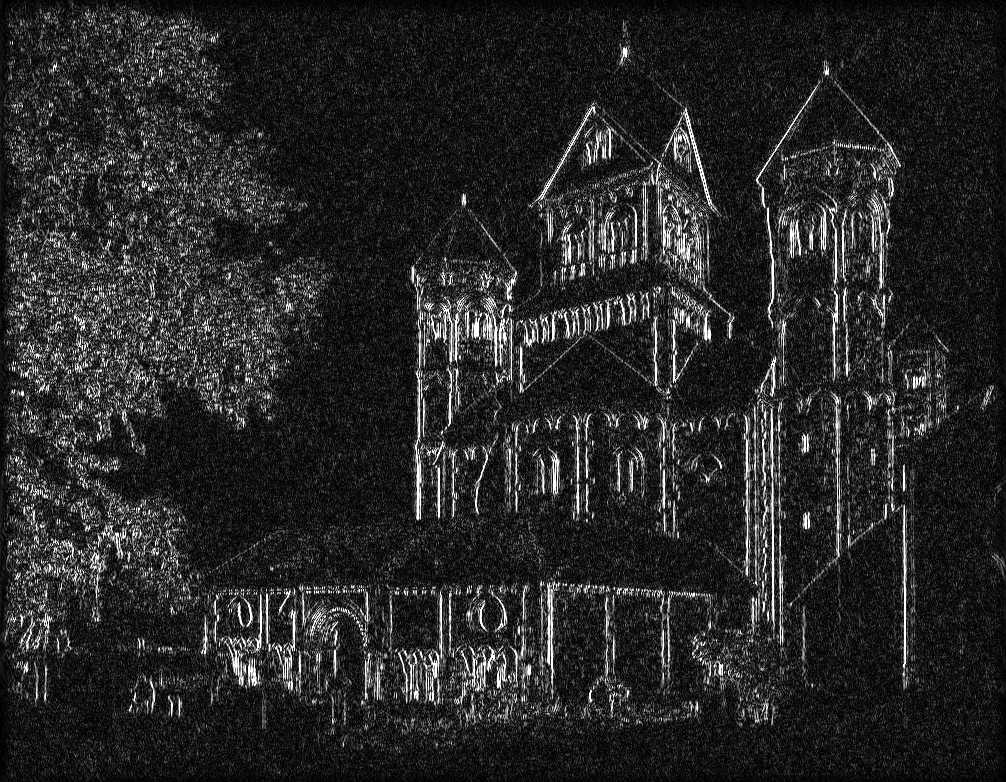}\\
				\end{tabular}
			\end{minipage}\\
			
			\begin{minipage}[b]{0.01\linewidth}
				\centering
				$\vcenter{\rotatebox{90}{L1 loss}}$
			\end{minipage}
			\vspace{-1mm}\begin{minipage}[t]{0.99\linewidth}
				\centering
				\begin{tabular}{cc}
					\includegraphics[width=0.45\linewidth]{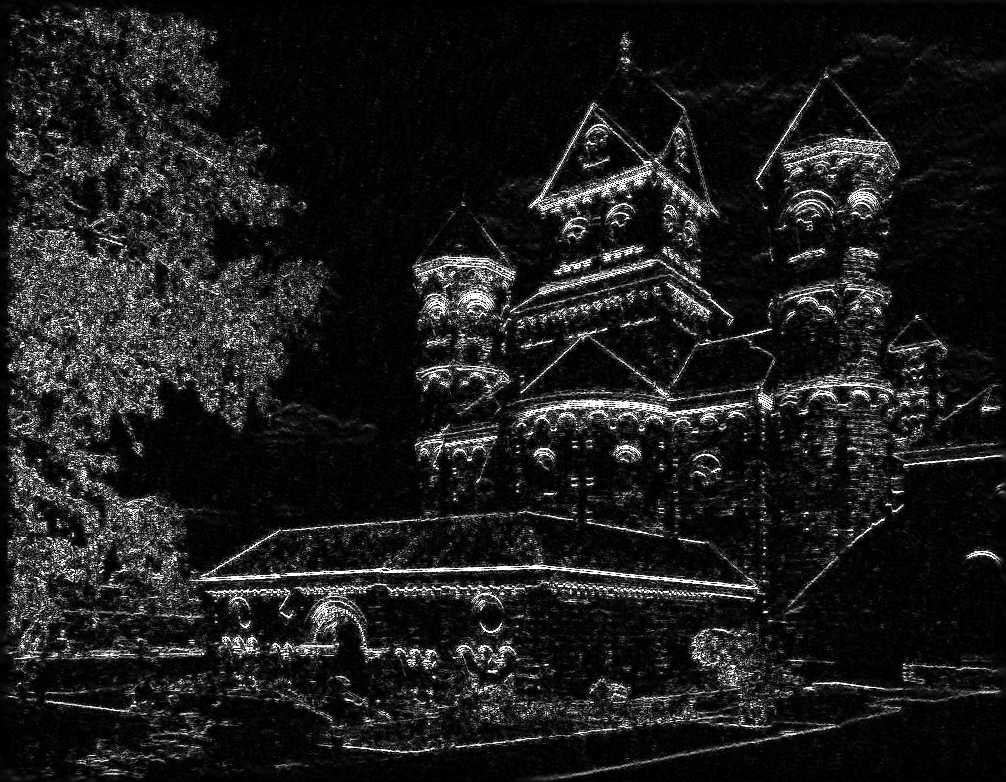}&
					\includegraphics[width=0.45\linewidth]{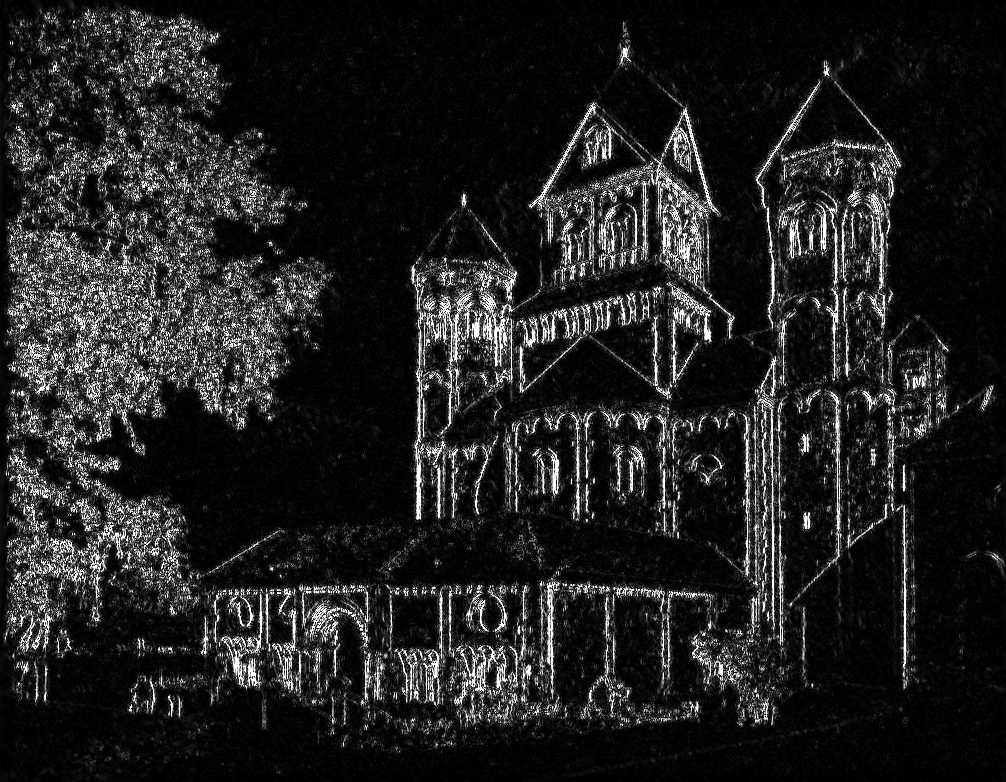}\\
				\end{tabular}
			\end{minipage}\\
			
			\begin{minipage}[b]{0.01\linewidth}
				\centering
				$\vcenter{\rotatebox{90}{Ground Truth}}$
			\end{minipage}
			\vspace{-1mm}\begin{minipage}[t]{0.99\linewidth}
				\centering
				\begin{tabular}{cc}
					\includegraphics[width=0.45\linewidth]{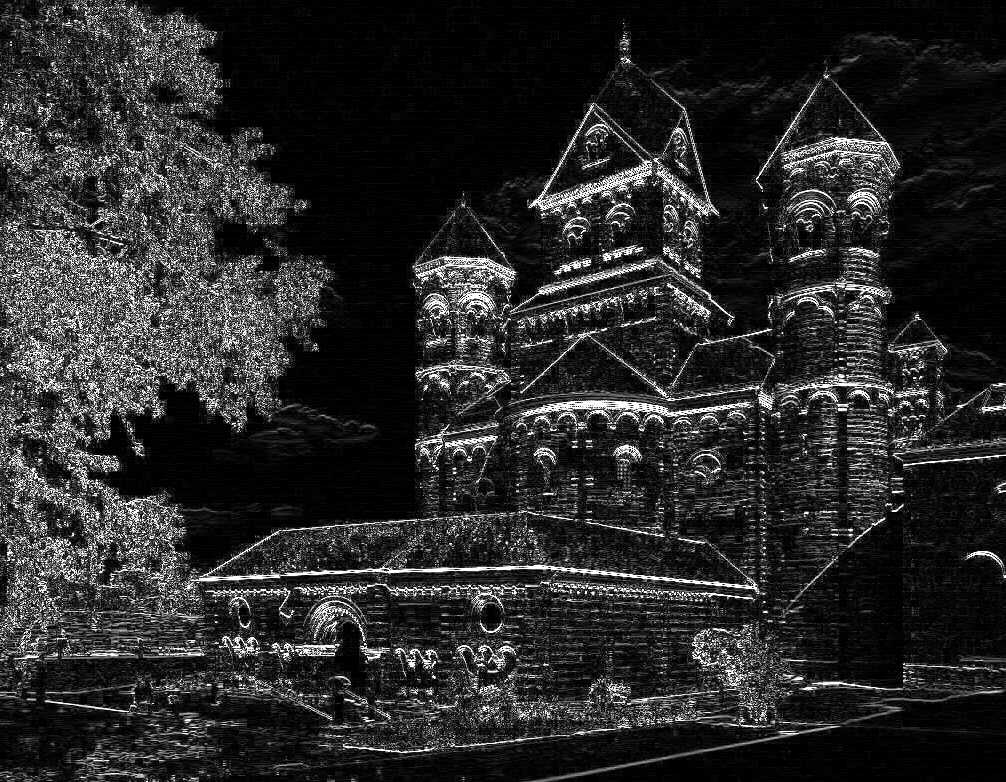}&
					\includegraphics[width=0.45\linewidth]{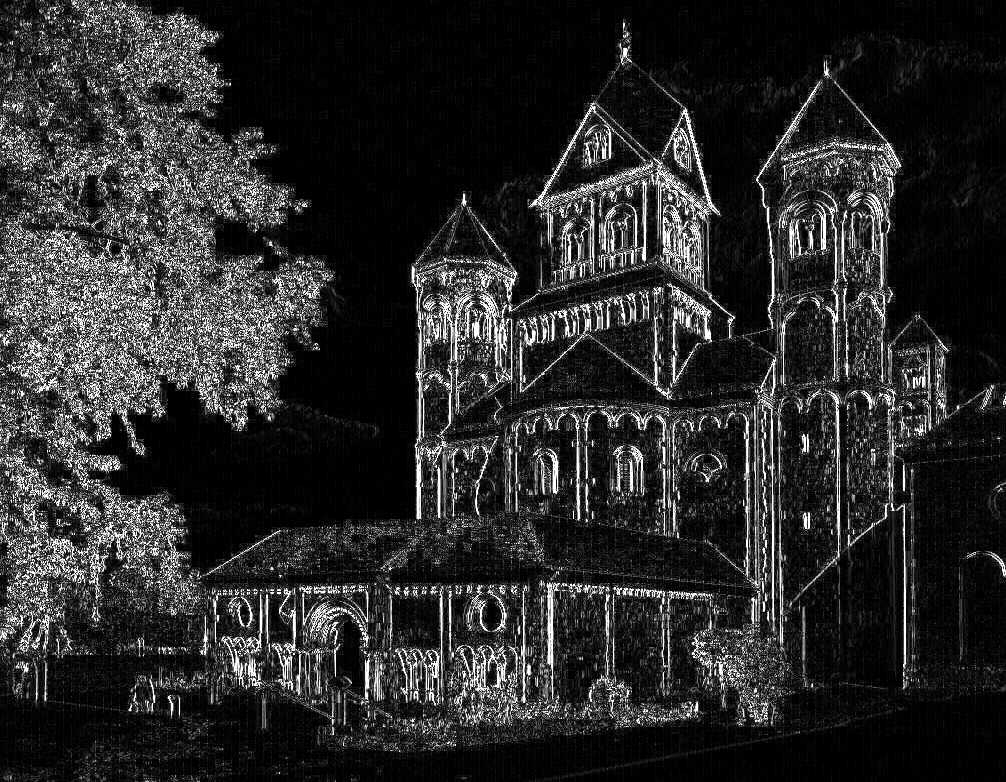}\\
					Vertical Gradient&Horizontal Gradient\\
				\end{tabular}
			\end{minipage}\\
		\end{tabular}
	\end{center}
	\caption{Visual comparison of gradient generation under different loss functions. The input blurry image is with 1\% noise. The vertical and horizontal gradient trained using L2 loss and L1 loss are shown on the first and second row, respectively. The ground truth gradient is shown on the last row. The gradient noise can be effectively reduced using L1 loss.}
	\label{fig:L1vsL2}
\end{figure}
\subsection{Effect of the Loss Function}
Most of existing CNN based low-level vision methods use the $L_2$ norm based reconstruction error as the loss function~\cite{dong2014learning}.
However, the $L_2$ norm is not robust to outliers and usually leads to results contains noise and ringing artifacts.
To overcome the limitations of $L_2$ norm based reconstruction error, we use an $L_1$ norm based reconstruction error as the loss function, \ie,~\eqref{eq: L1Norm}.

To validate the effect of the proposed loss function, we train the proposed network with the $L_2$ norm based reconstruction error and the $L_1$ norm based reconstruction error
using the same settings for the first iteration.
As shown in Figure~\ref{fig:L1vsL2_Loss}, the method with the $L_1$ norm based reconstruction error converges better
than that of the $L_2$ norm based reconstruction error.
Figure~\ref{fig:L1vsL2} shows that using  $L_1$ norm based reconstruction error is able to remove noise and artifacts compared to that of $L_2$ norm based reconstruction error.


\section{Experimental Results}
We evaluate the proposed algorithm against the state-of-the-art methods using benchmark datasets for non-blind deconvolution.
All the MATLAB source code and datasets will be made available to the public. 
More results can be found in the supplementary material.

\subsection{Training}

{\flushleft \bf{Parameter settings.}}
To train the network, we optimize the hyper-parameters and the weights of FCNN iteratively.
Specifically, the weights of FCNN are updated iteration by iteration with fixed hyper-parameters and then hyper-parameters are trained in an end-to-end manner with fixed weights of FCNN .
%
We implement the proposed algorithm using the MatConvNet~\cite{vedaldi15matconvnet} package.
We use Xavier initialization for the FCNN weights of each layer.
The initial value in the hyper-parameters training stage is randomly initialized.
(But keeps the later iteration has smaller hyper-parameter than the former one.)
Stochastic gradient descent (SGD) is used to train the network.
The learning rate in the training of FCNN is 0.01.
For the learning rate in the hyper-parameter training stage, we set it to be 10 in the last deconvolution module and 10,000 in other modules.
The momentum value for both FCNN and hyper-parameters training is set to be 0.95.
Since hyper-parameters are easily stuck into local minimal.
We train the network with several hyper-parameter initializations and select the best one.
%

\begin{figure}[!t]
	\begin{center}
		\begin{tabular}{cccccc}
			\includegraphics[width=\swsix]{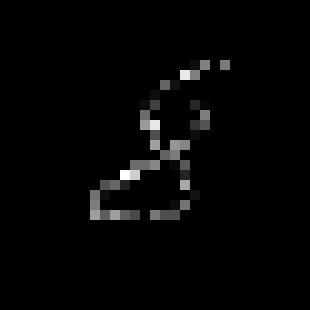} &
			\includegraphics[width=\swsix]{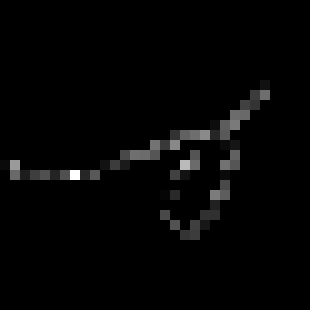} &
			\includegraphics[width=\swsix]{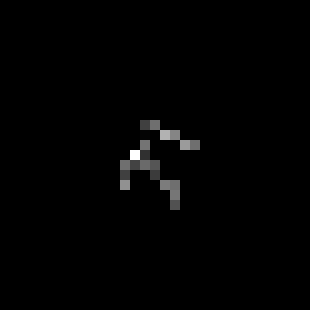} &
			\includegraphics[width=\swsix]{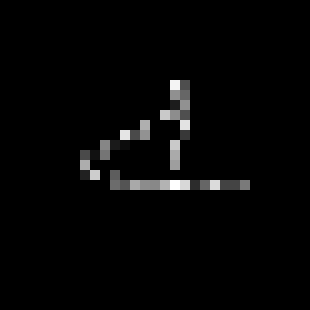} &
			\includegraphics[width=\swsix]{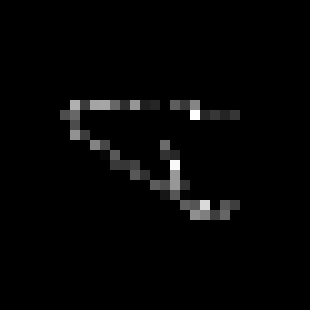} &
			\includegraphics[width=\swsix]{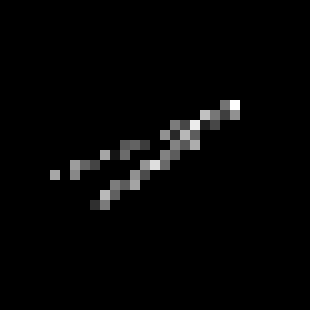} \\
		\end{tabular}
	\end{center}
	\caption{Examples of randomly generated kernels for training.}
	\label{fig:randKernels}
\end{figure}

{\flushleft \bf{Training dataset.}}
In order to generate enough blurred images for training, we use BSD500 dataset \cite{setbenchmarks}.
and randomly cropped image patches with a size of $256\times256$ pixels as the clear images.
%
The blurred kernels are generated according to \cite{chakrabarti2016neural}, whose size ranges from 11 to 31 pixels.
We generate blurred kernels according to \cite{chakrabarti2016neural}, where the size of blur kernels ranges from 11 to 31 pixels.
Some examples of our randomly generated kernels are shown in Figure~\ref{fig:randKernels}.
After obtaining these generated blur kernels, we convolve the clear image patches with blur kernels and plus Gaussian noises to obtain the blurry image patches.
And we train three networks with 1\%, 3\% and 5\% noises respectively.

\renewcommand{\tabcolsep}{.1pt}

{\flushleft \bf{Test dataset.}}
For the test dataset, we use the 80 ground truth clear images from the dataset by Sun~\etal~\cite{sun2013edge}
and eight blur kernels from the dataset by Levin~\etal~\cite{levin2009understanding}. Thus, we have 640 blurred images in total.
We evaluate all the methods on the blurred images with different Gaussian noise level which includes 1\%, 3\% and 5\%.
In addition to using the ground truth kernels from the dataset by Levin~\etal~\cite{levin2009understanding},
we also use the estimated blur kernels from the state-of-the-art blind deblurring methods ~\cite{pan2016robust} to examine the effectiveness of the proposed method.

\subsection{Comparisons with the State-of-the-Arts}
%
We compare the proposed iterative FCNN with other non-blind deblurring algorithms including HL~\cite{krishnan2009fast}, EPLL~\cite{zoran2011learning}, MLP~\cite{schuler2013machine},
and CSF~\cite{schmidt2014shrinkage}.
For the proposed method, we also use the proposed algorithm with one iteration and three iterations for comparison.
For fairness, we use the online available implementation of these methods and tuned the parameters to generate the best possible results.

\begin{figure*}[t]
	\begin{center}
		\begin{tabular}{cc}
			\begin{minipage}[b]{0.01\linewidth}
				\centering
				$\vcenter{\rotatebox{90}{\normalsize 1\% noise}}$
			\end{minipage}
			\begin{minipage}[t]{0.99\linewidth}
				\centering
				\begin{tabular}{cccccc}
					\vspace{-1mm}\includegraphics[width=0.161\linewidth]{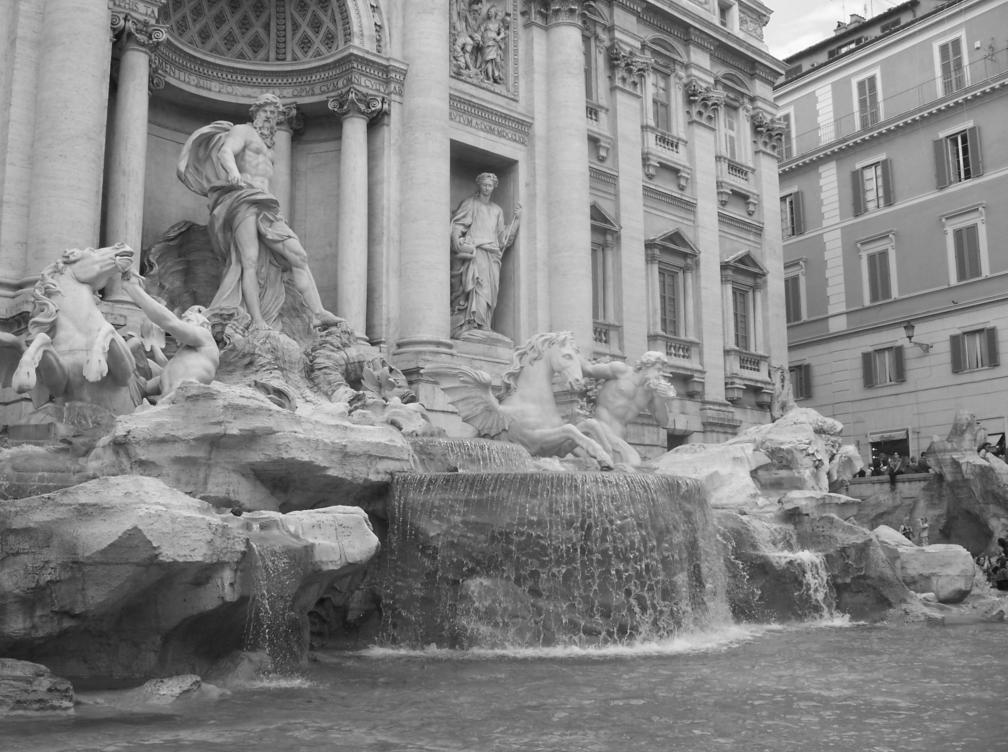}&
					\includegraphics[width=0.161\linewidth]{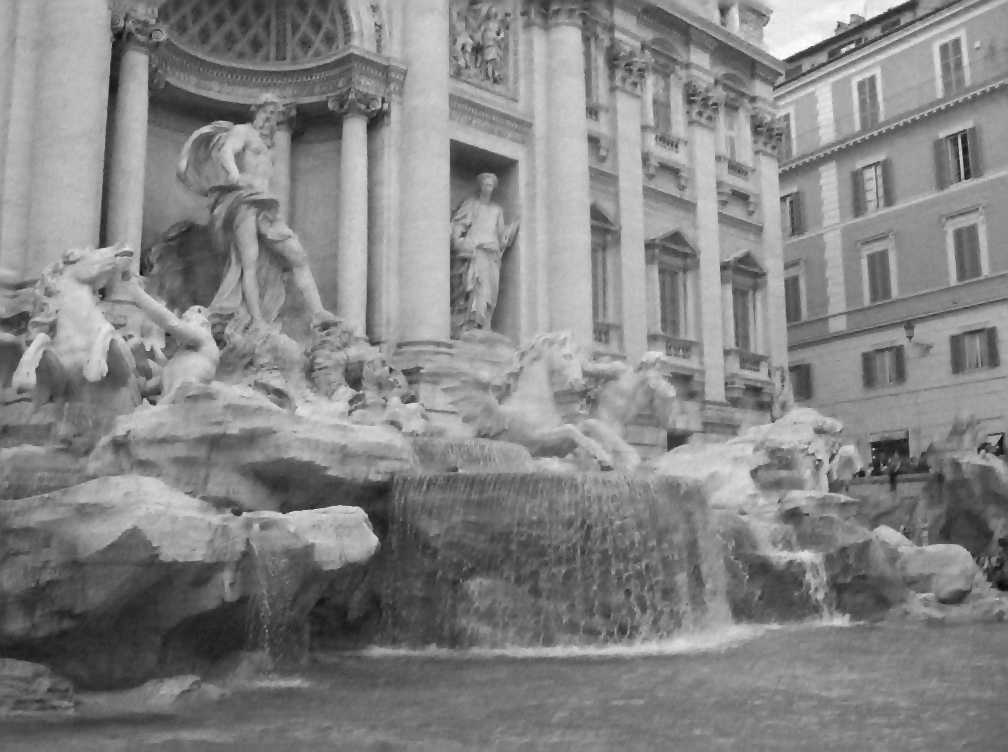}&
					\includegraphics[width=0.161\linewidth]{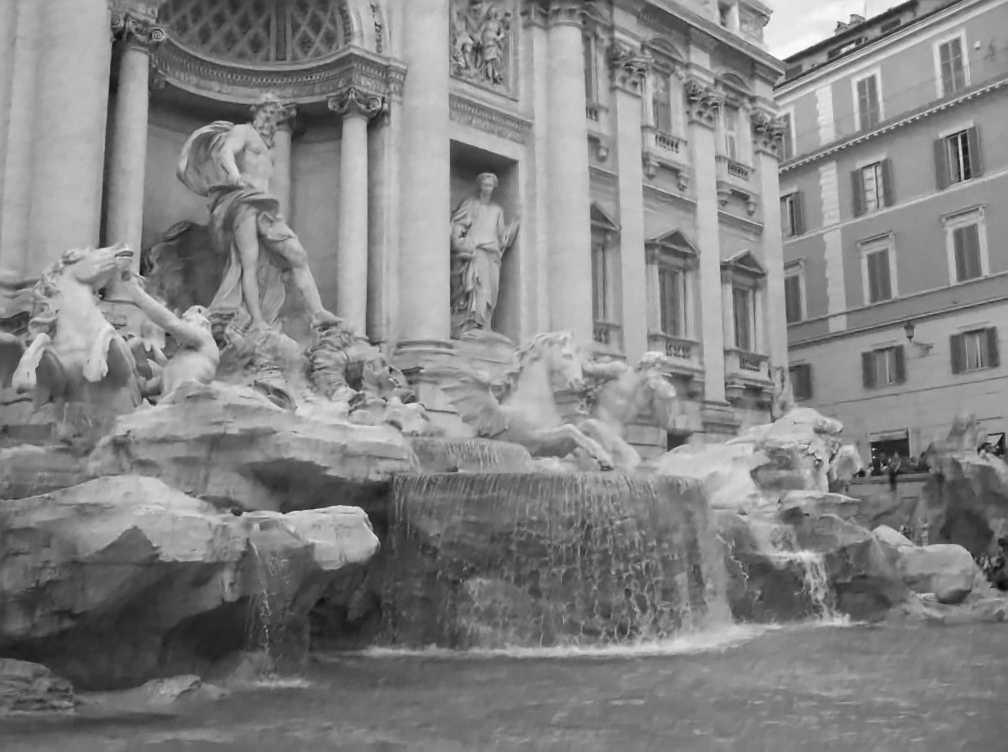}&
					\includegraphics[width=0.161\linewidth]{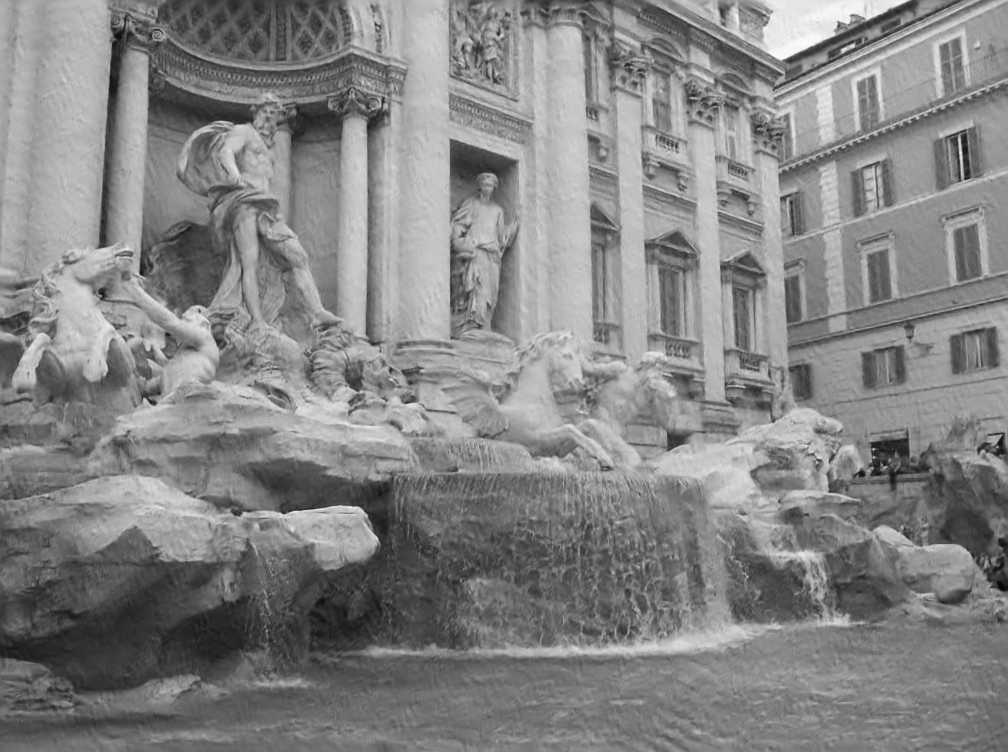}&
					\includegraphics[width=0.161\linewidth]{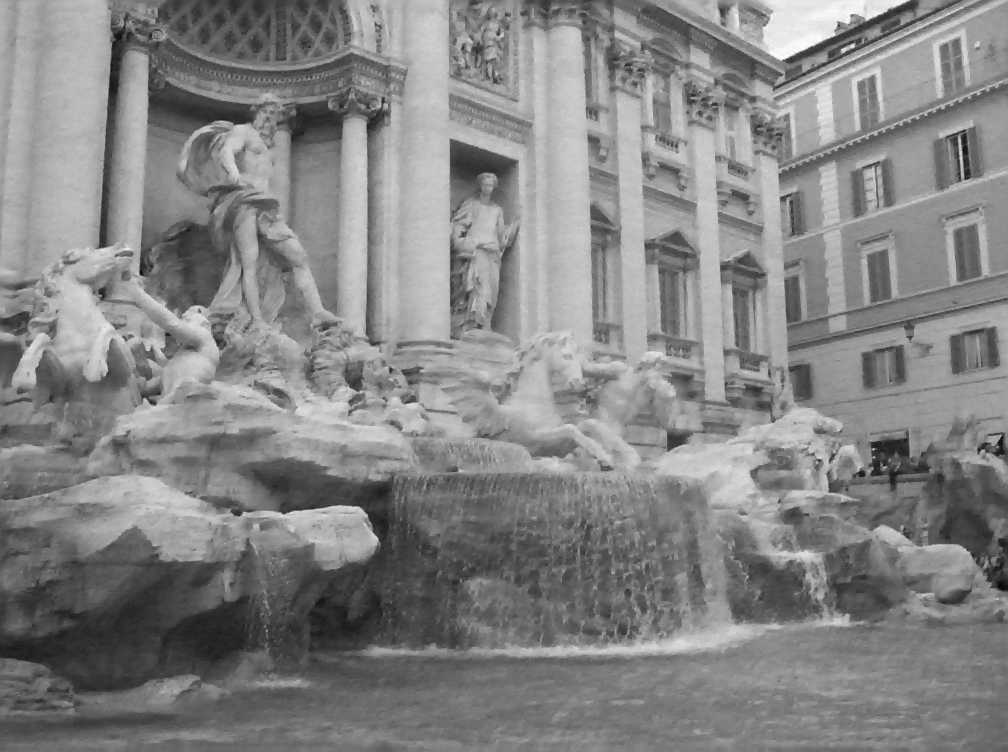}&
					\includegraphics[width=0.161\linewidth]{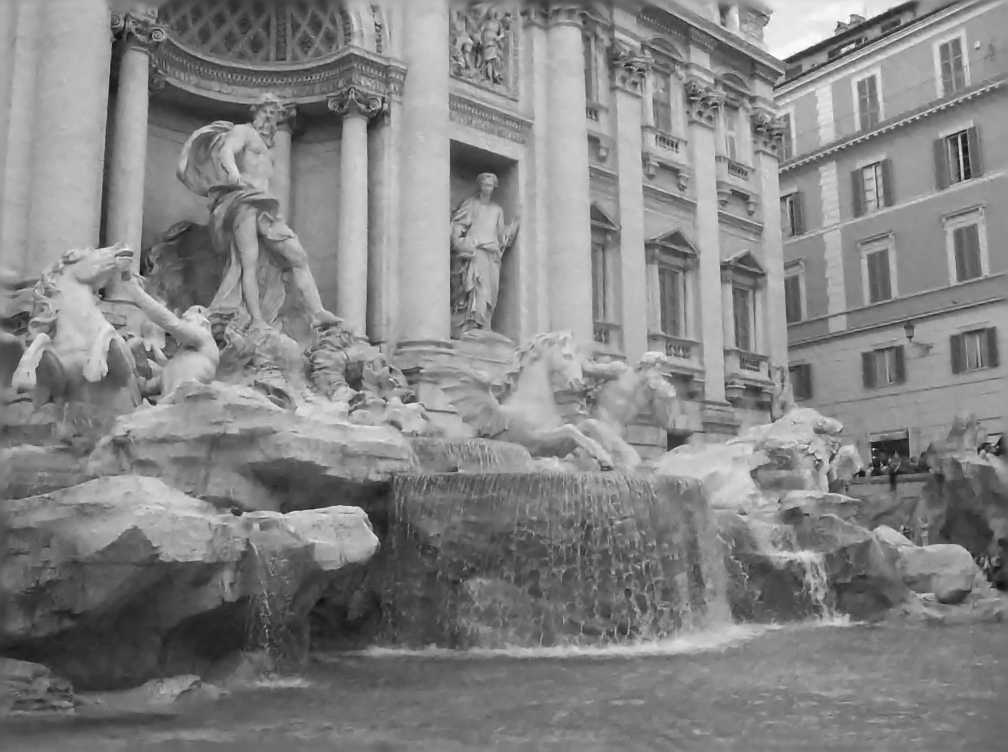}\\
					\vspace{-1mm}\includegraphics[width=0.161\linewidth]{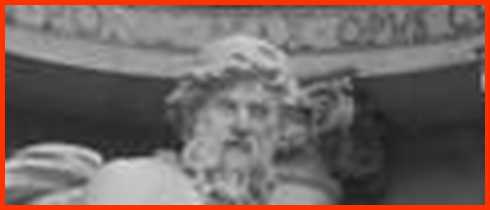}&
					\includegraphics[width=0.161\linewidth]{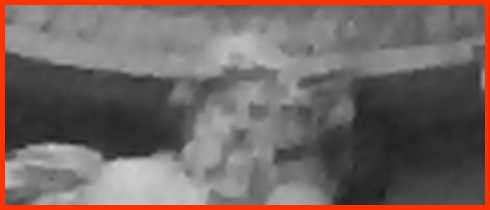}&
					\includegraphics[width=0.161\linewidth]{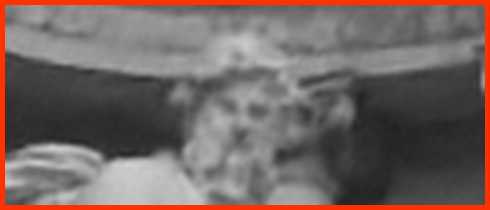}&
					\includegraphics[width=0.161\linewidth]{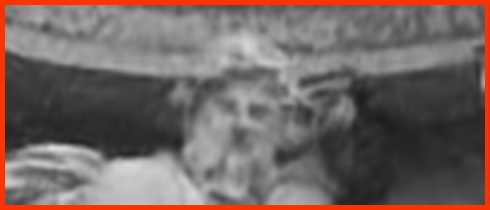}&
					\includegraphics[width=0.161\linewidth]{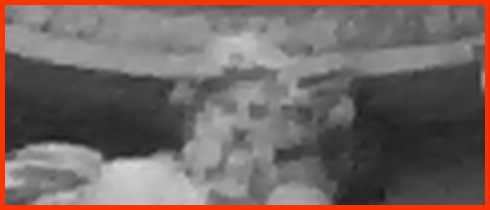}&
					\includegraphics[width=0.161\linewidth]{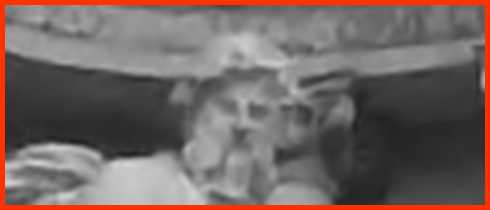}\\
					\footnotesize{PSNR / SSIM}&\footnotesize{30.45 / 0.84} & \footnotesize{32.05 / 0.88} & \footnotesize{31.20 / 0.85} & \footnotesize{30.89 / 0.85} & \footnotesize{\textbf{32.06} / \textbf{0.88}}\\
				\end{tabular}
			\end{minipage}\\
			
			\begin{minipage}[b]{0.01\linewidth}
				\centering
				$\vcenter{\rotatebox{90}{\normalsize 3\% noise}}$
			\end{minipage}
			\begin{minipage}[t]{0.99\linewidth}
				\centering
				\begin{tabular}{cccccc}
					\vspace{-1mm}\includegraphics[width=0.161\linewidth]{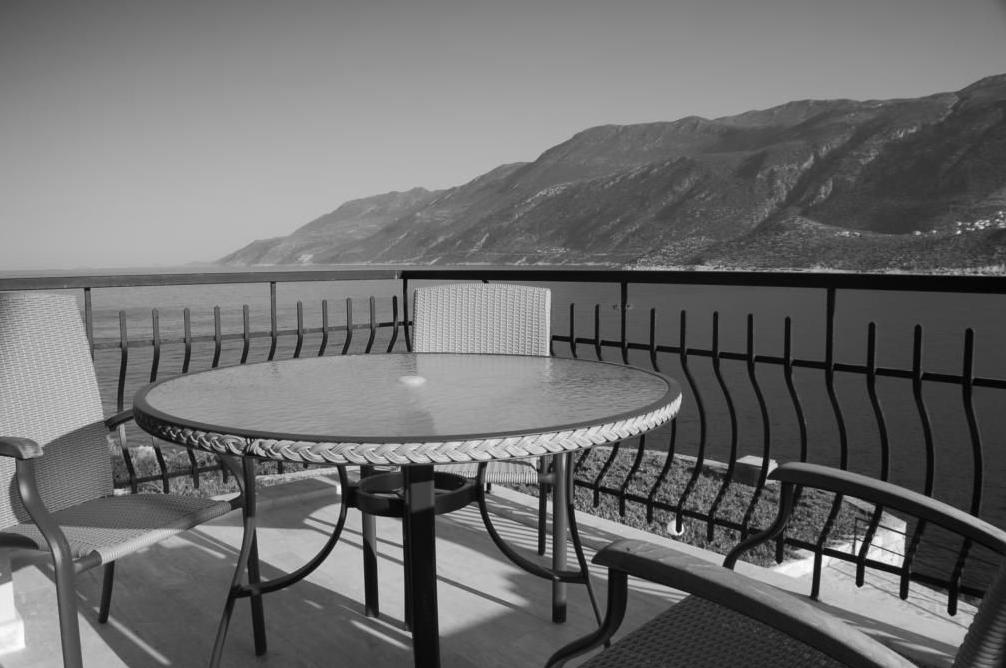}&
					\includegraphics[width=0.161\linewidth]{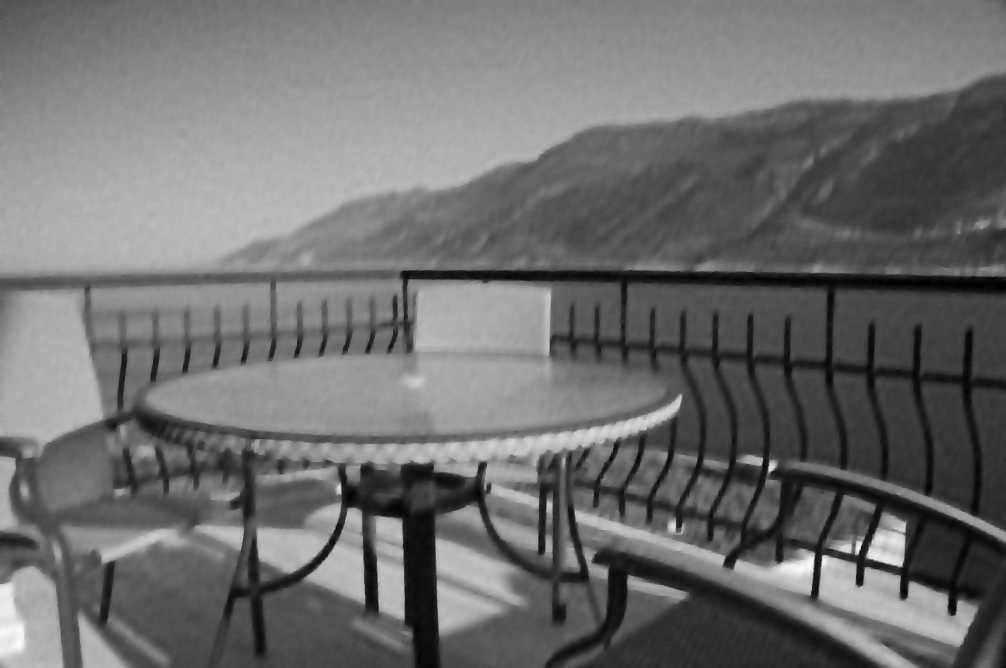}&
					\includegraphics[width=0.161\linewidth]{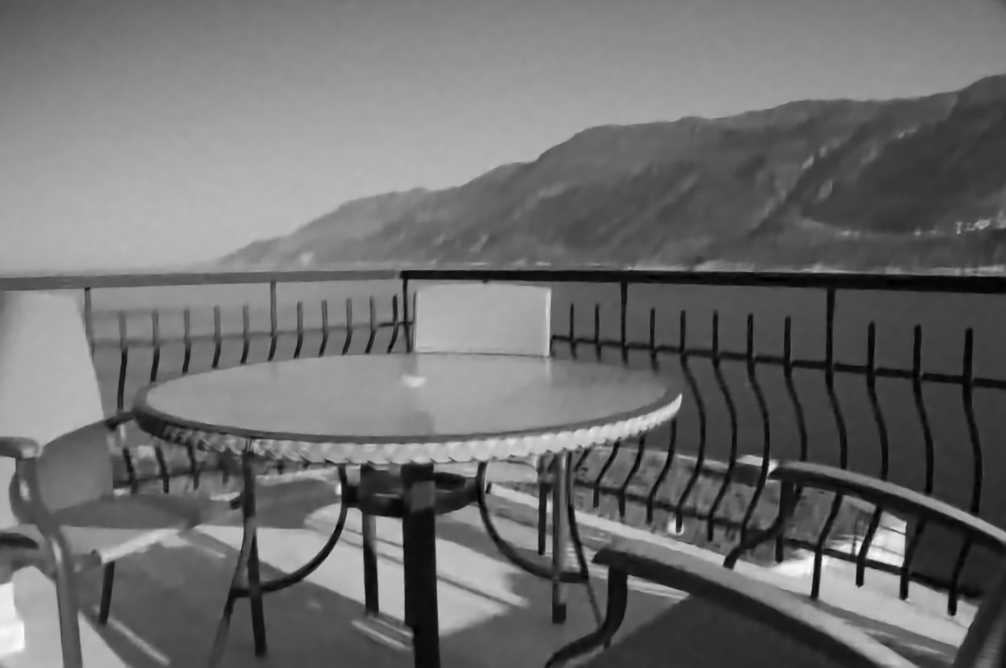}&
					\includegraphics[width=0.161\linewidth]{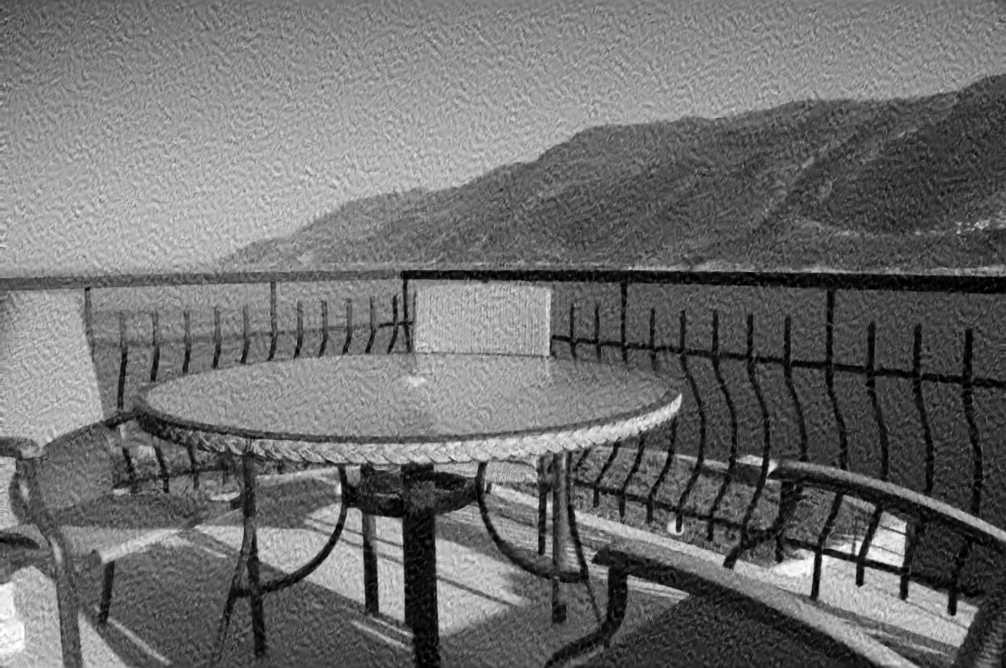}&
					\includegraphics[width=0.161\linewidth]{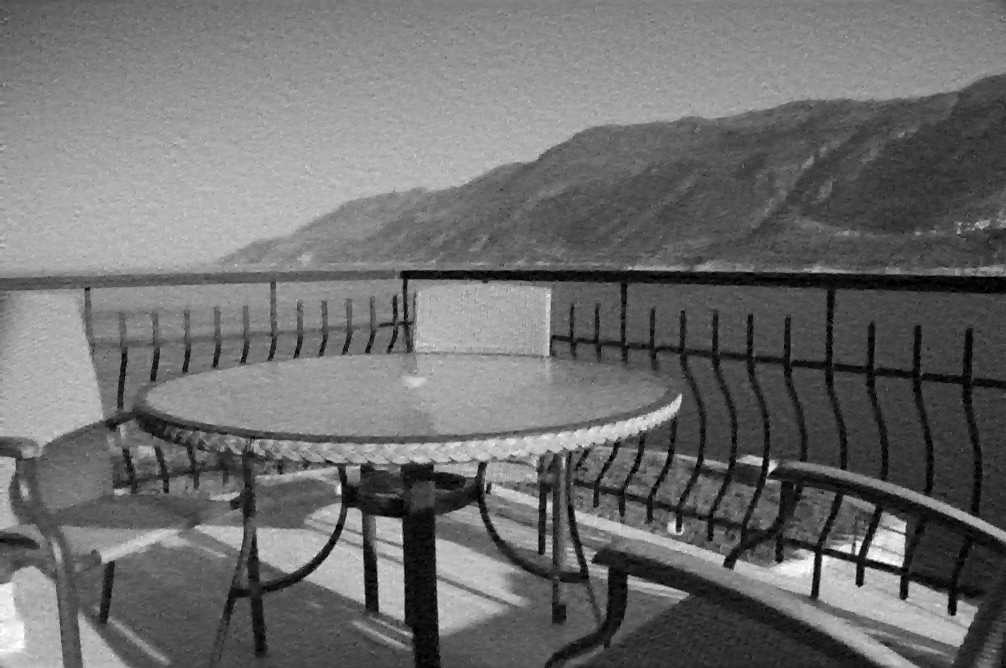}&
					\includegraphics[width=0.161\linewidth]{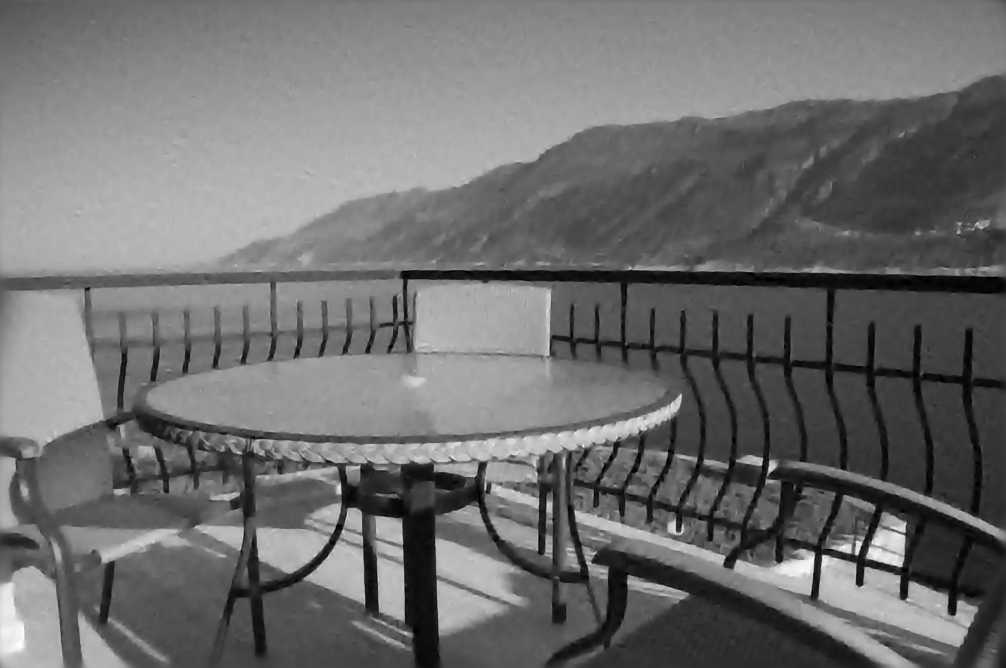}\\
					\vspace{-1mm}\includegraphics[width=0.161\linewidth]{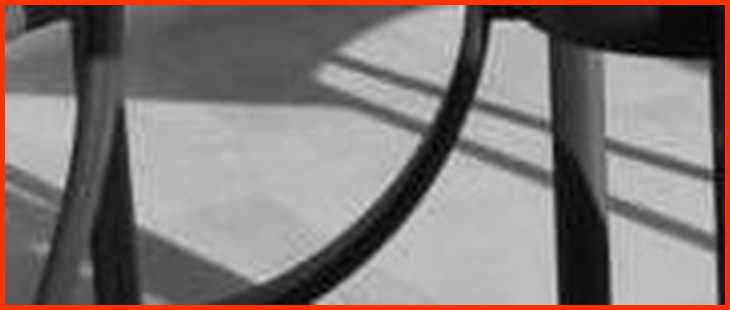}&
					\includegraphics[width=0.161\linewidth]{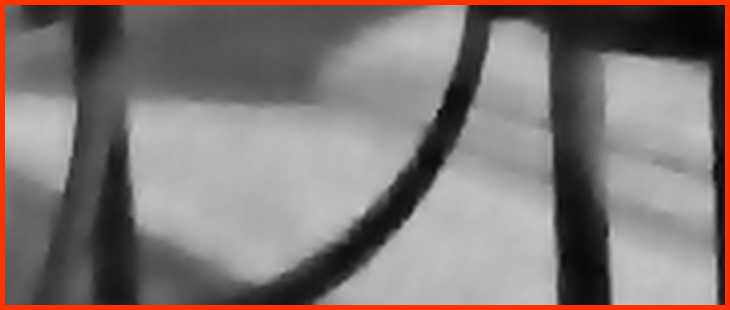}&
					\includegraphics[width=0.161\linewidth]{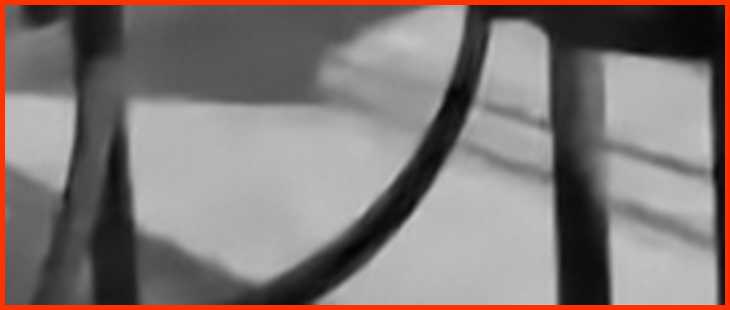}&
					\includegraphics[width=0.161\linewidth]{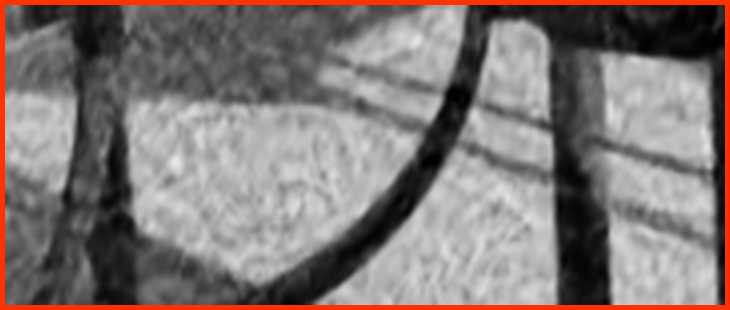}&
					\includegraphics[width=0.161\linewidth]{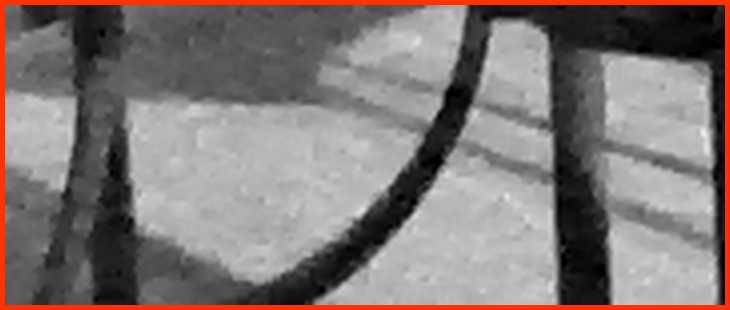}&
					\includegraphics[width=0.161\linewidth]{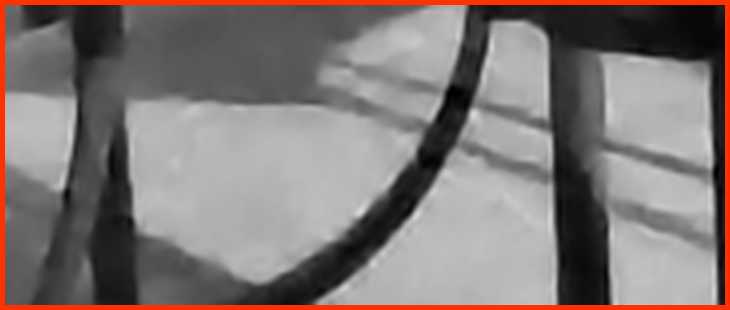}\\
					\footnotesize{PSNR / SSIM}&\footnotesize{27.58 / 0.79} & \footnotesize{29.34 / 0.84} & \footnotesize{25.57 / 0.51} & \footnotesize{28.66 / 0.75} & \footnotesize{\textbf{29.85} / \textbf{0.84}}\\
				\end{tabular}
			\end{minipage}\\
			
			\begin{minipage}[b]{0.01\linewidth}
				\centering
				$\vcenter{\rotatebox{90}{\normalsize 5\% noise}}$
			\end{minipage}
			\begin{minipage}[t]{0.99\linewidth}
				\centering
				\begin{tabular}{cccccc}
					\vspace{-1mm}\includegraphics[width=0.161\linewidth]{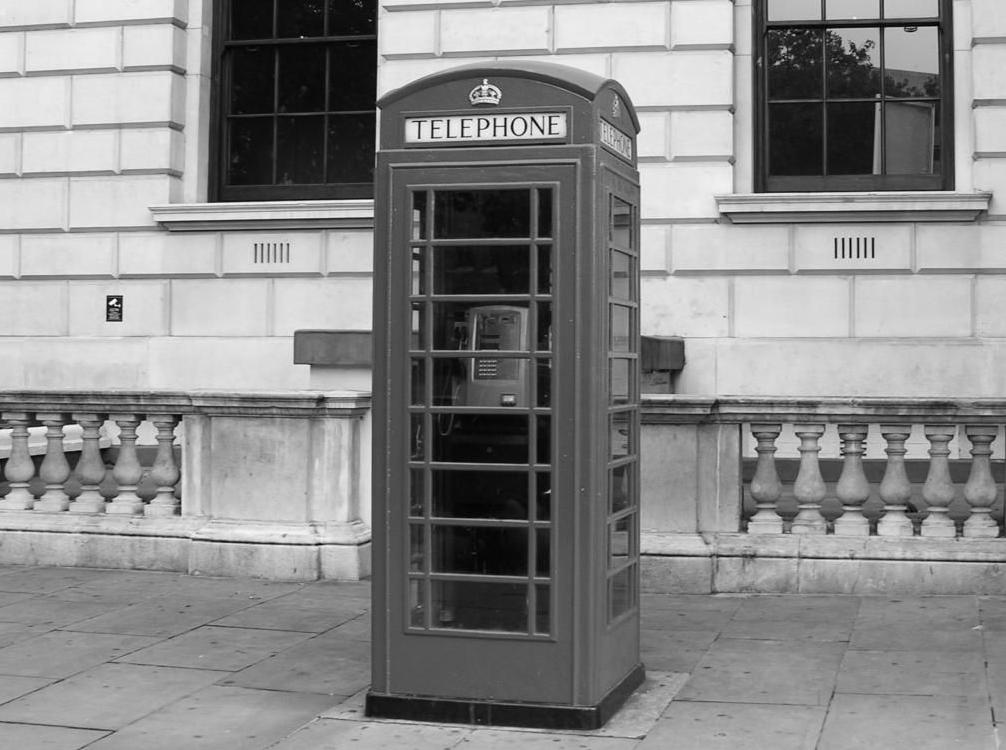}&
					\includegraphics[width=0.161\linewidth]{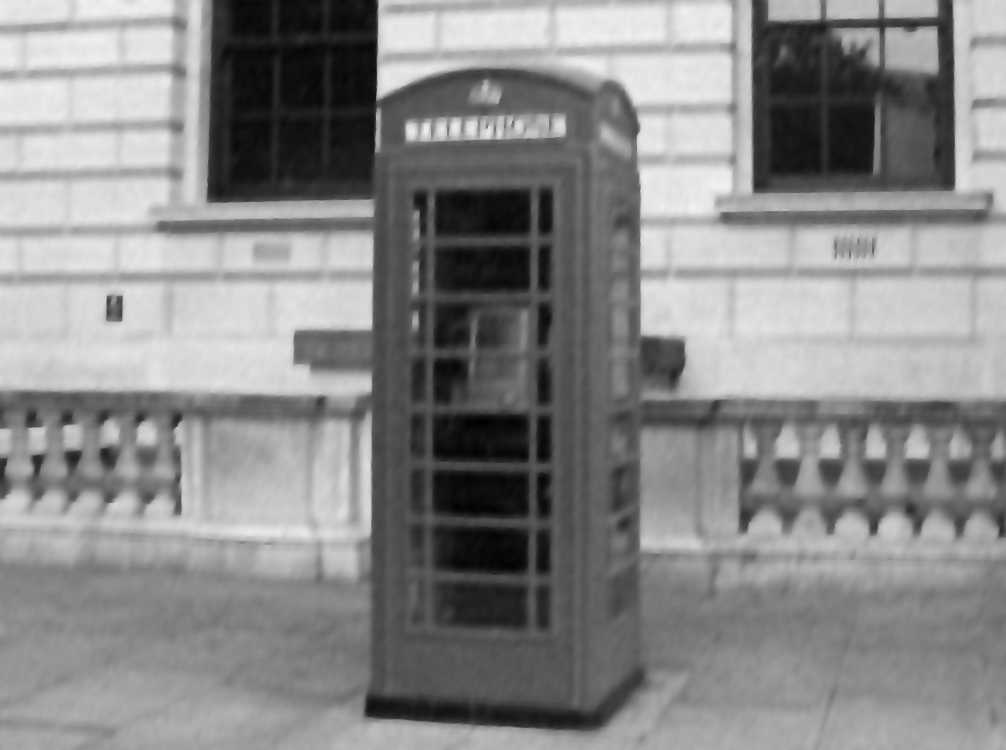}&
					\includegraphics[width=0.161\linewidth]{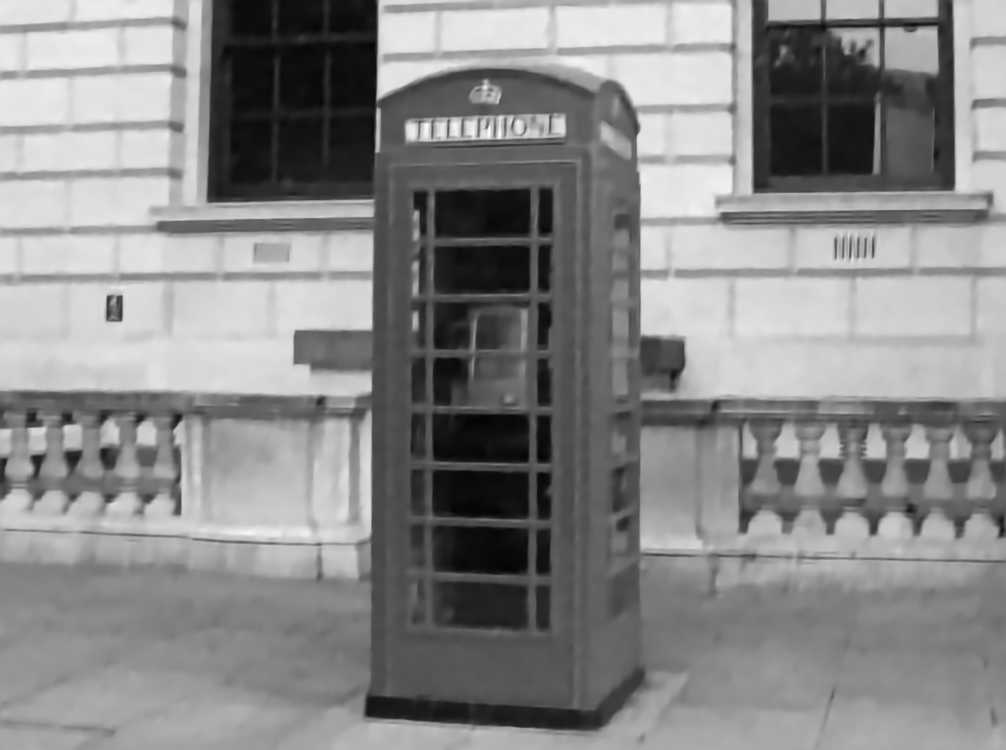}&
					\includegraphics[width=0.161\linewidth]{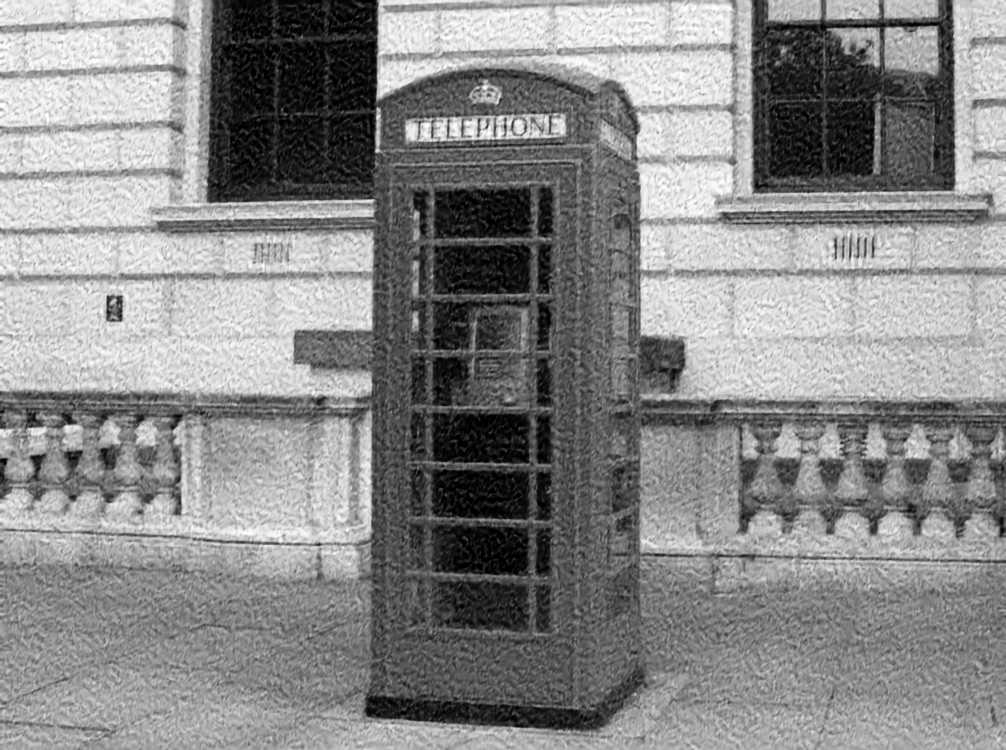}&
					\includegraphics[width=0.161\linewidth]{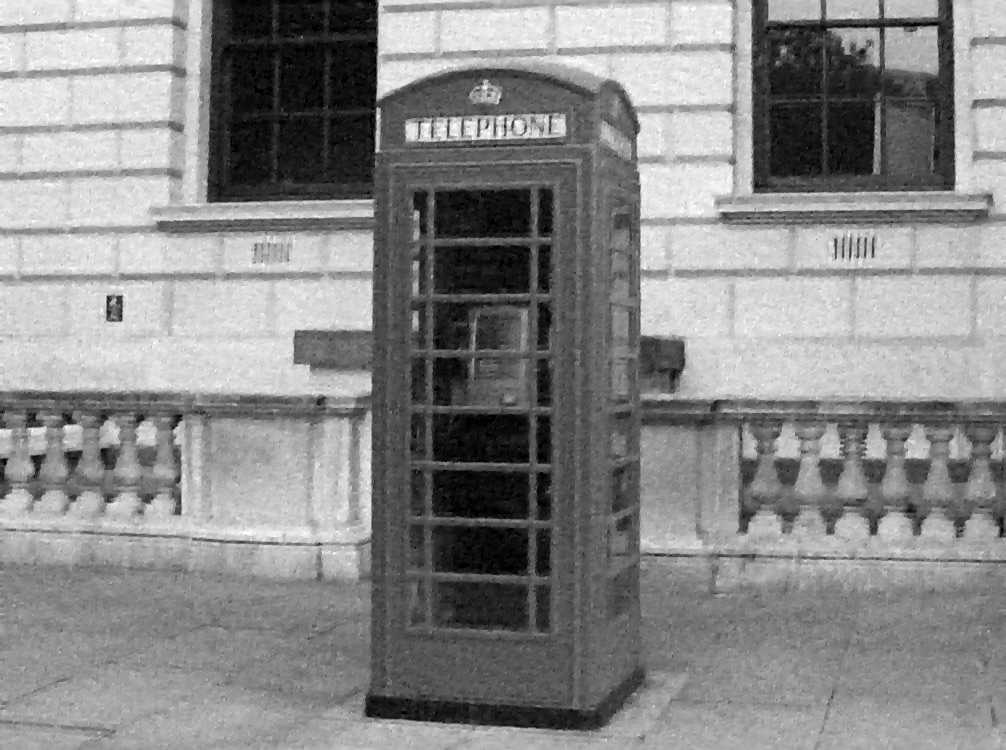}&
					\includegraphics[width=0.161\linewidth]{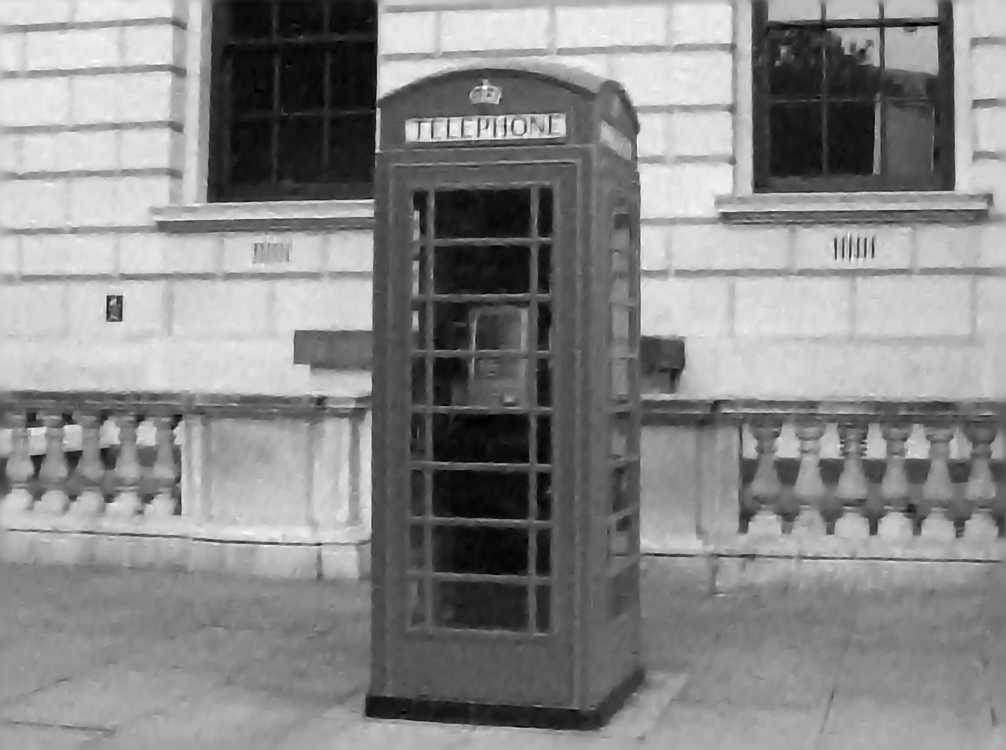}\\
					\vspace{-1mm}\includegraphics[width=0.161\linewidth]{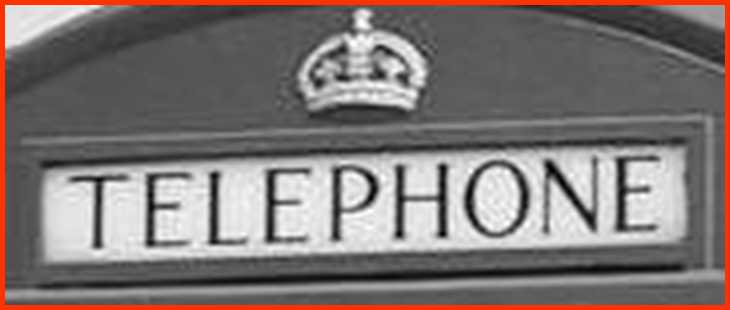}&
					\includegraphics[width=0.161\linewidth]{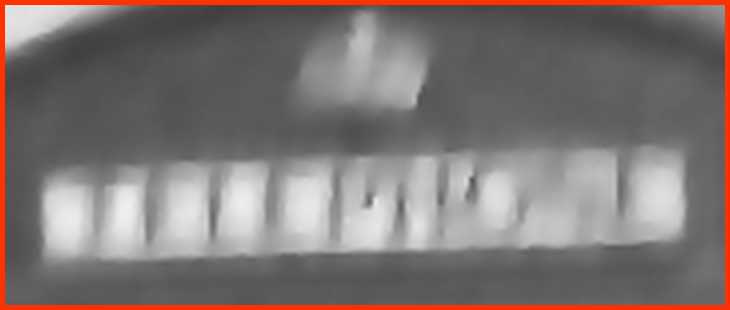}&
					\includegraphics[width=0.161\linewidth]{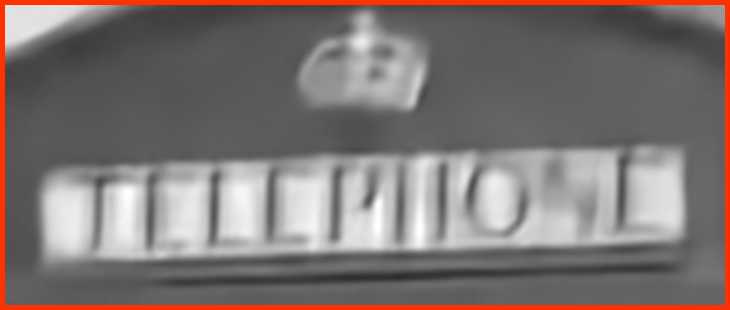}&
					\includegraphics[width=0.161\linewidth]{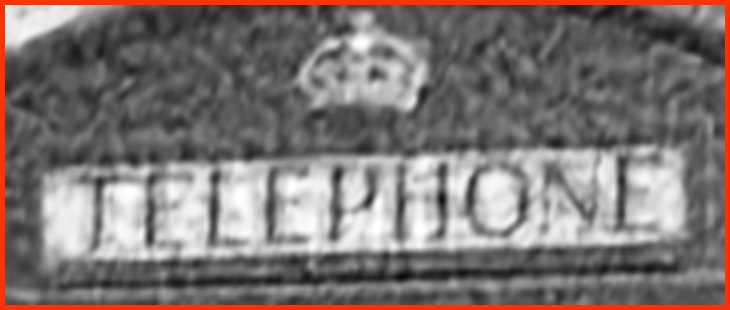}&
					\includegraphics[width=0.161\linewidth]{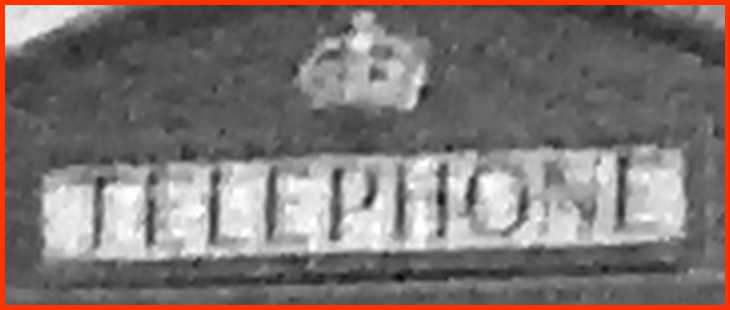}&
					\includegraphics[width=0.161\linewidth]{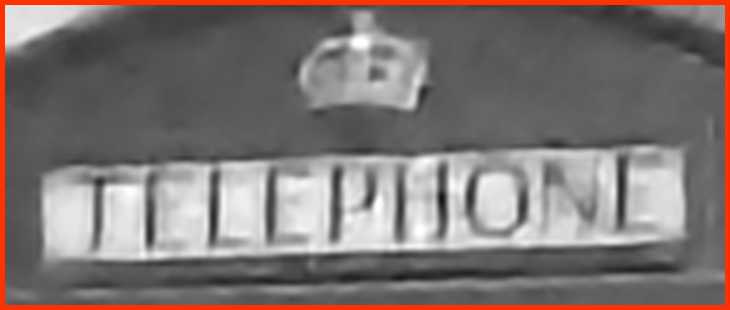}\\
					\vspace{-1mm}\footnotesize{PSNR / SSIM}&\footnotesize{24.16 / 0.71} & \footnotesize{26.04 / 0.77} & \footnotesize{23.91 / 0.48} & \footnotesize{26.09 / 0.67} & \footnotesize{\textbf{27.30} / \textbf{0.69}}\\
					\footnotesize{(a) Ground Truth} &\footnotesize{(b) HL\cite{krishnan2009fast}}&\footnotesize{(c) EPLL\cite{zoran2011learning}}&\footnotesize{(d) MLP\cite{schuler2013machine}}&\footnotesize{(e) CSF\cite{schmidt2014shrinkage}}&\footnotesize{(f) 3 iterations network}\\
				\end{tabular}
			\end{minipage}\\
		\end{tabular}
	\end{center}
	\caption{Visual evaluation under different input noise levels. The proposed method performs favorable compared with existing non-blind deblurring methods.}
	\label{fig:diffMethods}
\end{figure*}

We first quantitatively evaluate our method on the dataset with 1\% Gaussian noise and use PSNR and SSIM as the metrics.
As shown in Table~\ref{table:psnr_1}, our method outperforms HL~\cite{krishnan2009fast}, MLP~\cite{schuler2013machine} and CSF~\cite{schmidt2014shrinkage}
in terms of PSNR and SSIM metrics.
Although EPLL method performs slightly better than the proposed methods, this method is not efficient as it needs to solve complex optimization problems.
Furthermore, this method usually smooths details as shown in Figure~\ref{fig:diffMethods}(c), while our method generates the results with much clearer textures (Figure~\ref{fig:diffMethods}(f)).
We further note that the PSNR and SSIM values of the proposed iterative FCNN are higher than those of the proposed method with only one iteration network,
which demonstrates the effectiveness of the iterative FCNN method.
In addition, we use the estimated blur kernels from Pan\cite{pan2016robust} to evaluate the proposed method.
The PSNR and SSIM values in Table~\ref{table:psnr_1} demonstrate that the proposed method still performs well and can be applied to methods Pan\cite{pan2016robust}
to improve the performance of restored results.

We further evaluate our method on the images with 3\% and 5\% Gaussian noise. Tables~\ref{table:psnr_3} and \ref{table:psnr_5} show the results by different methods.
Our methods achieves better performance compared to HL~\cite{krishnan2009fast}, MLP~\cite{schuler2013machine} and CSF~\cite{schmidt2014shrinkage} when the noise level is high.
In addition to PSNR and SSIM, our method also generates much clearer images with fine textures as shown in Figure~\ref{fig:diffMethods}.


{\flushleft \bf{Running time.}}
The proposed method performs more efficiently than other state-of-the-art methods. Table~\ref{table:time} summarizes
the average run time of representative methods with different image resolutions.
HL and EPLL run on an Intel Core i7 CPU and MLP, CSF and our method run on a Nvidia K40 GPU.

\renewcommand{\tabcolsep}{10pt}
\begin{table}[!ht]
	\centering
	\caption{Average PSNR and SSIM for 1\% noises}
	\vspace{+3mm}
	\label{table:psnr_1}
	\begin{tabular}{|c|c|c|c|}
		\hline
		blur kernel                     & ground truth & Pan\cite{pan2016robust} \\ \hline
		HL\cite{krishnan2009fast}       & 31.57/0.87   & 29.94/0.84              \\ \hline
		EPLL\cite{zoran2011learning}    & 33.00/0.89   & 30.61/0.87              \\ \hline
		MLP\cite{schuler2013machine}    & 31.82/0.86   & 28.76/0.80              \\ \hline
		CSF\cite{schmidt2014shrinkage}  & 31.93/0.87   & 30.22/0.86              \\ \hline
		1 iteration                     & 32.50/0.89   & 30.38/0.86              \\ \hline
		3 iterations                    & 32.82/0.90   & 30.39/0.87              \\ \hline
	\end{tabular}
\end{table}

\begin{table}[!ht]
	\centering
	\caption{Average PSNR and SSIM for 3\% noises}
	\vspace{+3mm}
	\label{table:psnr_3}
	\begin{tabular}{|c|c|c|c|}
		\hline
		blur kernel                     & ground truth & Pan\cite{pan2016robust} \\ \hline
		HL\cite{krishnan2009fast}       & 27.42/0.73   & 26.91/0.72              \\ \hline
		EPLL\cite{zoran2011learning}    & 28.71/0.78   & 27.75/0.77              \\ \hline
		MLP\cite{schuler2013machine}    & 26.26/0.60   & 25.04/0.57              \\ \hline
		CSF\cite{schmidt2014shrinkage}  & 28.43/0.78   & 27.11/0.74              \\ \hline
		1 iteration                     & 28.71/0.77   & 27.34/0.75              \\ \hline
		3 iterations                    & 29.12/0.79   & 27.58/0.77              \\ \hline
	\end{tabular}
\end{table}

\begin{table}[!ht]
	\centering
	\caption{Average PSNR and SSIM for 5\% noises}
	\vspace{+3mm}
	\label{table:psnr_5}
	\begin{tabular}{|c|c|c|c|}
		\hline
		blur kernel                     & ground truth & Pan\cite{pan2016robust} \\ \hline
		HL\cite{krishnan2009fast}       & 25.85/0.67   & 25.48/0.66              \\ \hline
		EPLL\cite{zoran2011learning}    & 27.00/0.71   & 26.24/0.71              \\ \hline
		MLP\cite{schuler2013machine}    & 24.62/0.51   & 22.32/0.45              \\ \hline
		CSF\cite{schmidt2014shrinkage}  & 26.92/0.67   & 24.86/0.65              \\ \hline
		1 iteration                     & 27.25/0.72   & 25.49/0.69              \\ \hline
		3 iterations                    & 27.66/0.74   & 25.77/0.72              \\ \hline
	\end{tabular}
\end{table}

\renewcommand{\tabcolsep}{5pt}
\begin{table}[!ht]
	\centering
	\caption{Average time cost (seconds) with different image sizes for three iterations network. HL and EPLL run on a Intel Core i7 CPU and MLP, CSF and our method run on a Nvidia K40 GPU.}
	\vspace{+3mm}
	\label{table:time}
	\begin{tabular}{|c|c|c|c|c|c|}
		\hline
		image size & HL & EPLL & MLP & CSF & ours \\
		& \cite{krishnan2009fast} & \cite{zoran2011learning} & \cite{schuler2013machine} & \cite{schmidt2014shrinkage} &  \\ \hline
		$512\times400$     & 0.31				   & 209.58    				  & 0.80 					 & 0.08							  & 0.02 \\ \hline
		$1024\times800$    & 0.71				   & 953.52					  & 2.98 					 & 0.09							  & 0.03 \\ \hline
		$1536\times1200$   & 2.11				   & N/A					  & 6.73 					 & 0.33							  & 0.06 \\ \hline
	\end{tabular}
\end{table}

\renewcommand{\tabcolsep}{0.1pt}
\begin{figure*}[!htb]
	\begin{center}
		\begin{tabular}{ccc}
			\includegraphics[width=\swthree]{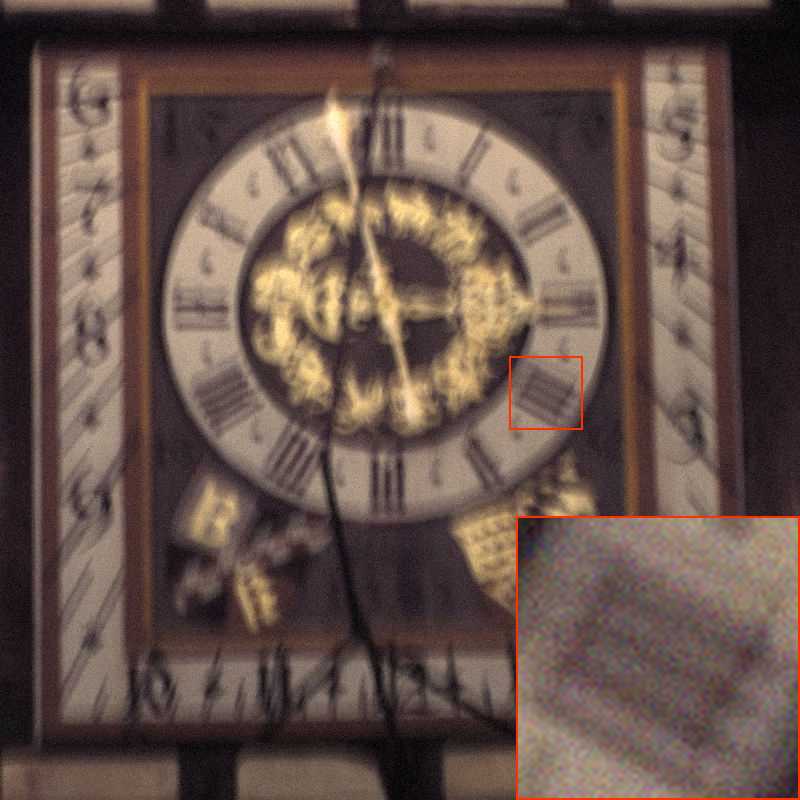} &
			\includegraphics[width=\swthree]{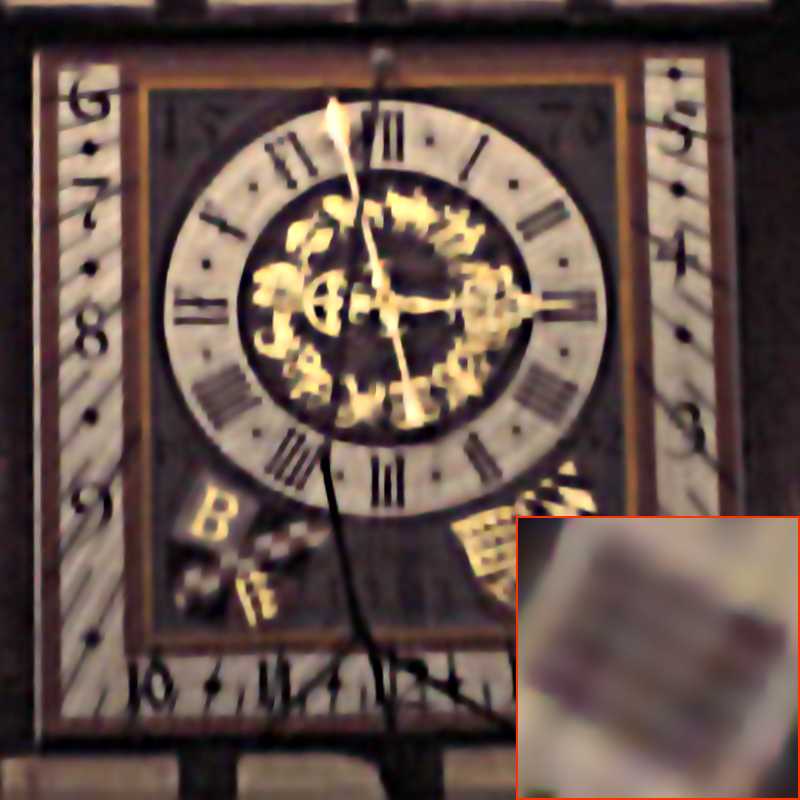} &
			\includegraphics[width=\swthree]{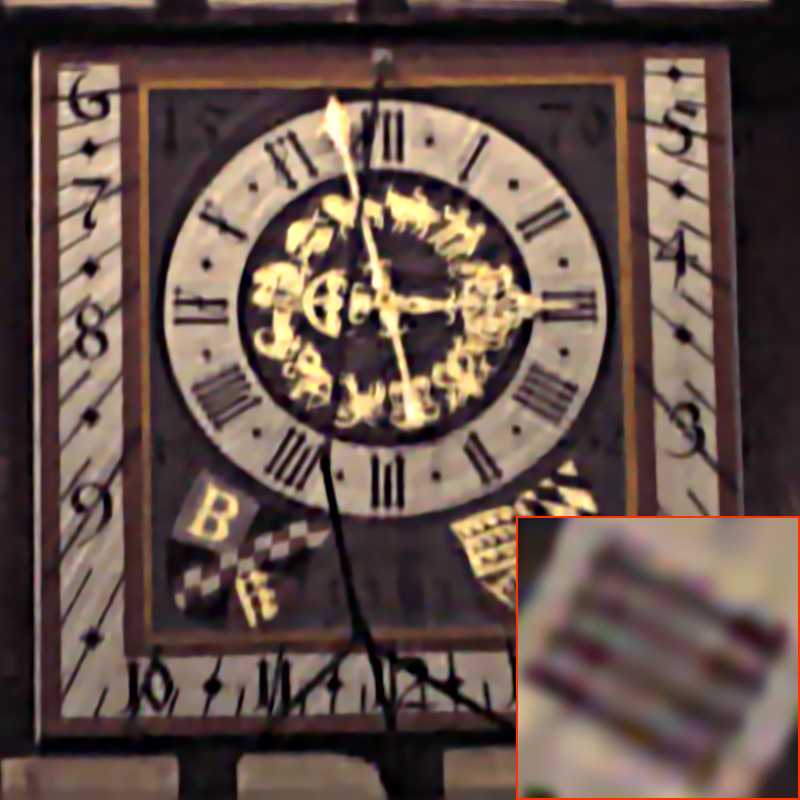} \\
			(a) blurry image &(b) HL\cite{krishnan2009fast}&(c) EPLL\cite{zoran2011learning}\\
			\includegraphics[width=\swthree]{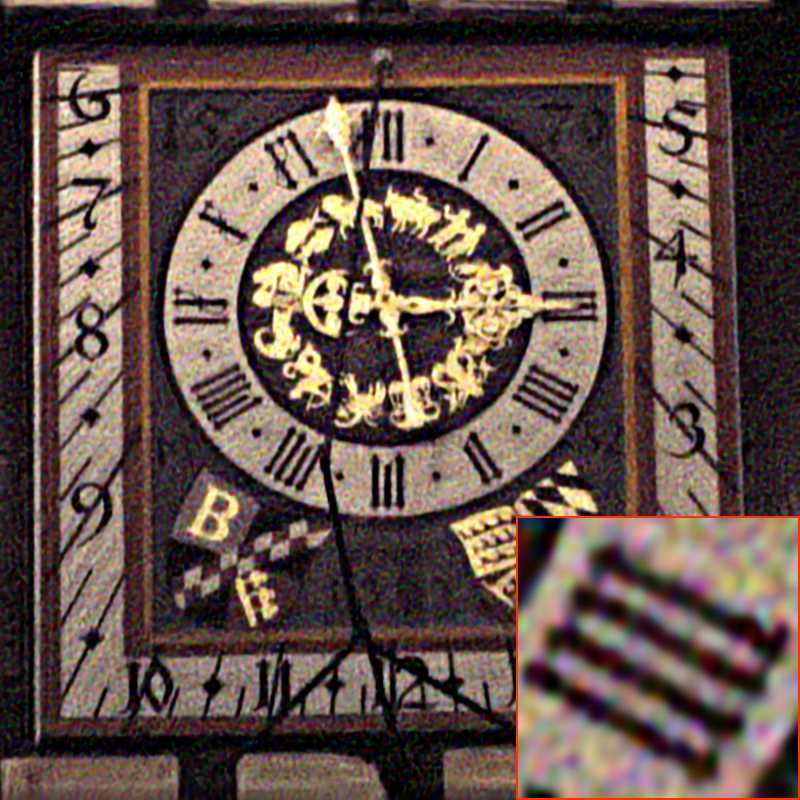} &
			\includegraphics[width=\swthree]{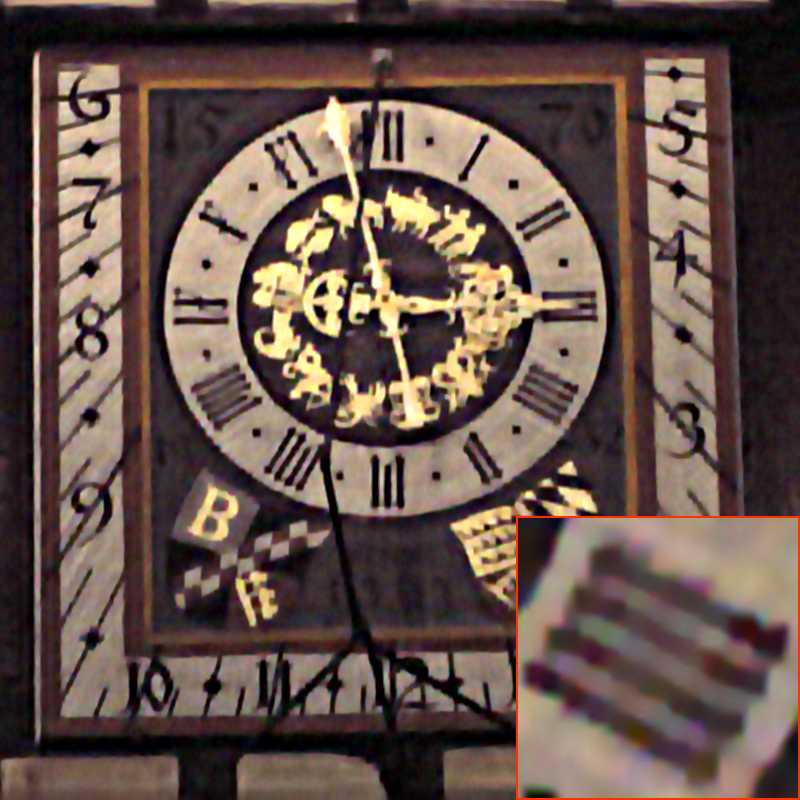} &
			\includegraphics[width=\swthree]{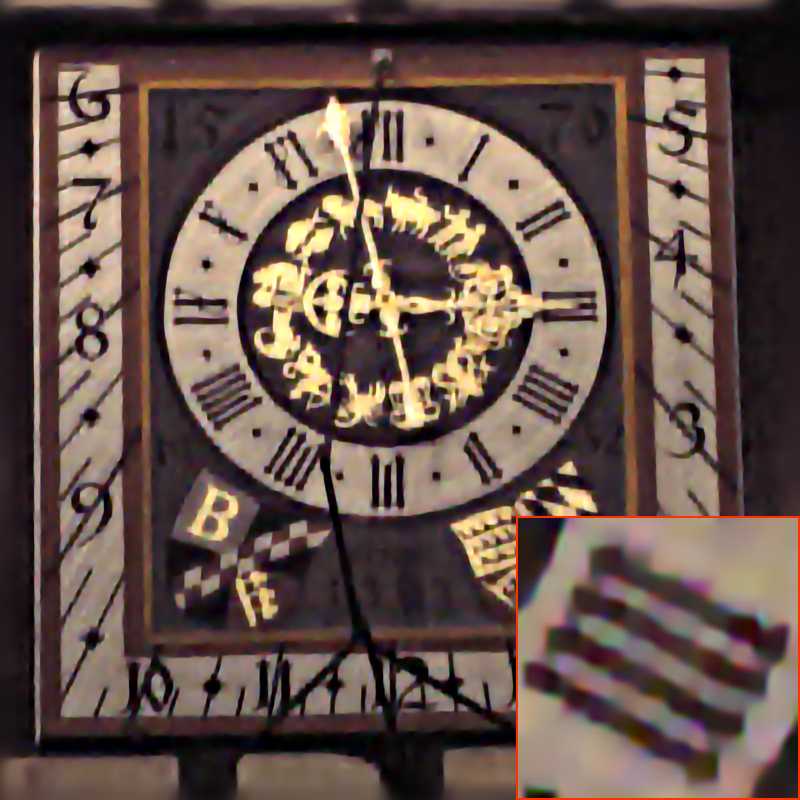} \\
			(d) MLP\cite{schuler2013machine}&(e) CSF\cite{schmidt2014shrinkage}&(f) 3 iterations network\\
		\end{tabular}
	\end{center}
	\caption{Visual evaluation for real world blurry image. The proposed method performs favorable compared with existing non-blind deblurring methods.}
	\label{fig:diffMethodsRealImg}
\end{figure*}

\subsection{Experiments for real blurry images}
We also test our three iterations network for one real blurry image from \cite{kohler2012recording}.
We add 3\% Gaussian noises into the original blurred image and use the network trained with this noise level for this experiment.
We use \cite{pan2016robust} to estimate kernel of the blurry image.
\reffig{fig:diffMethodsRealImg} shows that our three iterations network has comparable performance relative to EPLL. CSF cannot remove all the noise especially in flat regions and the result of HL is too blurry.

\section{Conclusions}

We propose an efficient non-blind deconvolution algorithm based on a fully convolutional neural network (FCNN).
The proposed method involves deconvolution part and denoising part, where the denoising part is achieved by a FCNN.
The learned features from FCNN is able to help the deconvolution.
To remove noises and ringing artifacts, we develop an FCNN for iterative deconvolution, which is able preserve image details.
Furthermore, we propose a hyper-parameters learning algorithm to improve the performance of image restoration.
Our method performs favorably against
state-of-the-art methods on both synthetic and real-world images.
Importantly, our approach is much more efficient
because FCNN is used for denoising part which is implemented in GPU in parallel.
{\small
\bibliographystyle{ieee}
\bibliography{cvpr17_CNN_deblur}
}

\end{document}